\documentclass[10pt,twocolumn,letterpaper]{article}

\usepackage{cvpr}              

\usepackage{multirow}
\usepackage{multicol}
\usepackage{listings}

\definecolor{cvprblue}{rgb}{0.21,0.49,0.74}
\usepackage[pagebackref,breaklinks,colorlinks,allcolors=cvprblue]{hyperref}

\def\paperID{xxx} 
\def\confName{CVPR}
\def\confYear{2026}

\title{Fast SceneScript: Fast and Accurate Language‑Based 3D Scene Understanding via Multi‑Token Prediction}

\author{
Ruihong Yin$^{1,2}$ \ Xuepeng Shi$^{1}$ \ Oleksandr Bailo$^{1}$ \ Marco Manfredi$^{1}$ \ Theo Gevers$^{2}$\\
$^1$Qualcomm XR Labs \ $^2$University of Amsterdam\\
}

\newcommand\blfootnote[1]{%
  \begingroup
  \renewcommand\thefootnote{}\footnote{#1}%
  \addtocounter{footnote}{-1}%
  \noindent
  \endgroup
}

\begin{document}
\maketitle
\begin{abstract}

Recent perception-generalist approaches based on language models have achieved state-of-the-art results across diverse tasks, including 3D scene layout estimation and 3D object detection, via unified architecture and interface. However, these approaches rely on autoregressive next-token prediction, which is inherently slow. In this work, we introduce Fast SceneScript, a novel structured language model for accurate and efficient 3D scene understanding. Our method employs multi-token prediction (MTP) to reduce the number of autoregressive iterations and significantly accelerate inference. While MTP improves speed, unreliable token predictions can significantly reduce accuracy. To filter out unreliable tokens, we adapt self-speculative decoding (SSD) for structured language models and introduce confidence-guided decoding (CGD) with an improved scoring mechanism for token reliability. Furthermore, we design a parameter-efficient mechanism that reduces the parameter overhead of MTP. Extensive experiments on synthetic and real-world benchmarks demonstrate that Fast SceneScript can generate up to 9 tokens per decoder inference step without compromising accuracy, while adding only $\sim7.5\%$ additional parameters.
\blfootnote{
Ruihong Yin completed the work during internship at Qualcomm.
}

\end{abstract}
\section{Introduction}
\label{sec:intro}

Recent success of language models~\cite{vaswani2017attention, DBLP:journals/corr/abs-2302-13971, DBLP:journals/corr/abs-2412-19437} has spawn a new wave of building a perception generalist model which can solve various perception problems with a single unified architecture and interface. Pix2Seq \cite{chenpix2seq,chen2022unified}, VisionLLM~\cite{DBLP:conf/nips/WangCCWZZLLZQD23}, and Rex-Omni~\cite{DBLP:journals/corr/abs-2510-12798} propose to use one language model to solve multiple 2D perception problems, such as object detection~\cite{DBLP:conf/nips/RenHGS15, DBLP:conf/eccv/CarionMSUKZ20}, instance segmentation~\cite{DBLP:conf/iccv/HeGDG17,DBLP:conf/cvpr/0004GGH18}, keypoint detection~\cite{DBLP:conf/iccv/HeGDG17,DBLP:conf/cvpr/0004GGH18}, and image captioning~\cite{vaswani2017attention,DBLP:conf/cvpr/PanYLM20}. SceneScript~\cite{avetisyan2024SceneScript, DBLP:journals/corr/abs-2503-11806} and SpatialLM~\cite{mao2025spatiallm} are language-based generalist models which can achieve state-of-the-art accuracy in 3D perceptions tasks, such as 3D scene layout estimation~\cite{yue2023connecting}, 3D object detection~\cite{DBLP:conf/cvpr/Brazil0SR0G23,DBLP:conf/iccv/MisraGJ21,DBLP:conf/wacv/RukhovichV022}, and coarse 3D object part reconstruction~\cite{DBLP:conf/cvpr/TulsianiSGEM17,DBLP:journals/tog/YangC21}.

\begin{figure}[t]
  \centering
  \includegraphics[width=1.0\linewidth]{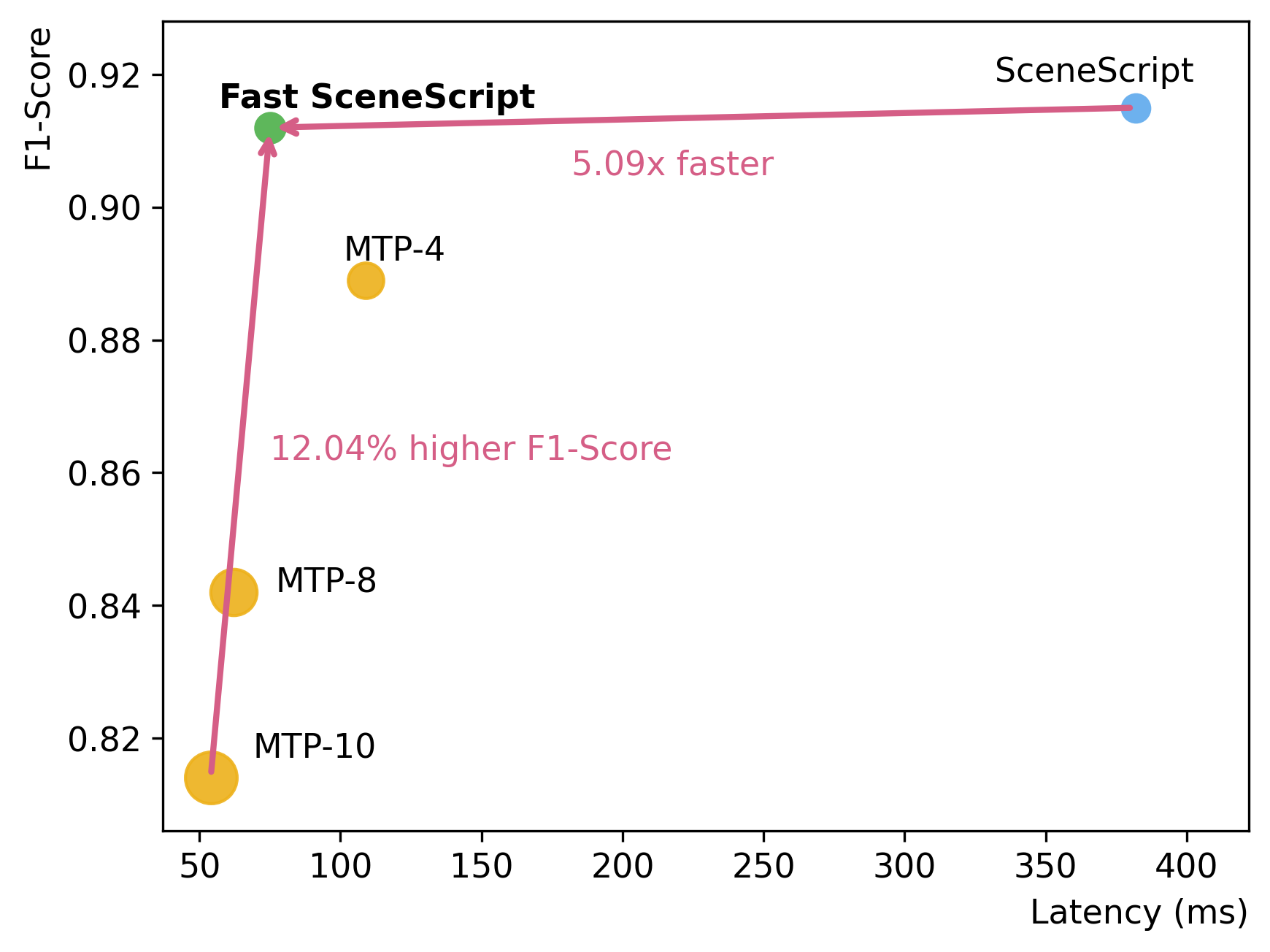}

   \caption{Comparison of SceneScript~\cite{avetisyan2024SceneScript}, SceneScript~\cite{avetisyan2024SceneScript} + MTP~\cite{gloeckle2024better}, and Fast SceneScript for layout estimation. Circle area is proportional to the number of decoder parameters. MTP-$n$ denotes SceneScript~\cite{avetisyan2024SceneScript} with $n$ token heads predicting $n$ future tokens. Fast SceneScript exhibits markedly higher computational efficiency than SceneScript~\cite{avetisyan2024SceneScript}, achieving a $5.09\times$ speedup. Additionally, our method delivers a $12.04\%$ improvement in F1-Score while utilizing $43\%$ fewer parameters compared to MTP-$n$.}
   \label{fig:teaser}
\end{figure}

\begin{figure*}[t]
  \centering
  \begin{subfigure}[t]{0.175\linewidth}
    \centering
    \includegraphics[width=\linewidth]{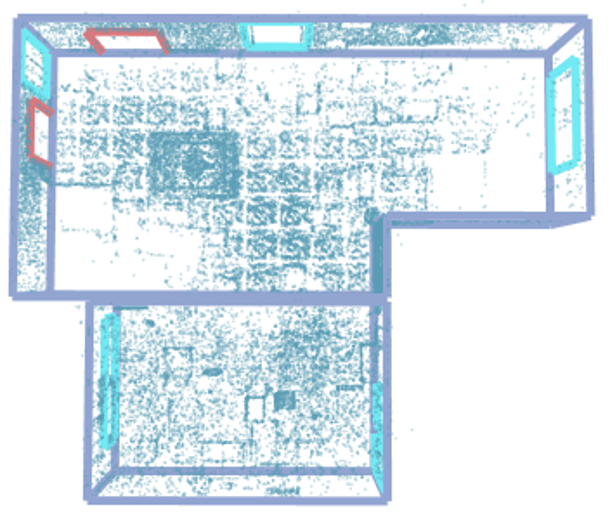}
    \caption{3D scene layout}
    \label{fig:mtp_concept_a}
  \end{subfigure}\hfill
  \begin{subfigure}[t]{0.385\linewidth}
    \centering
    \includegraphics[width=\linewidth]{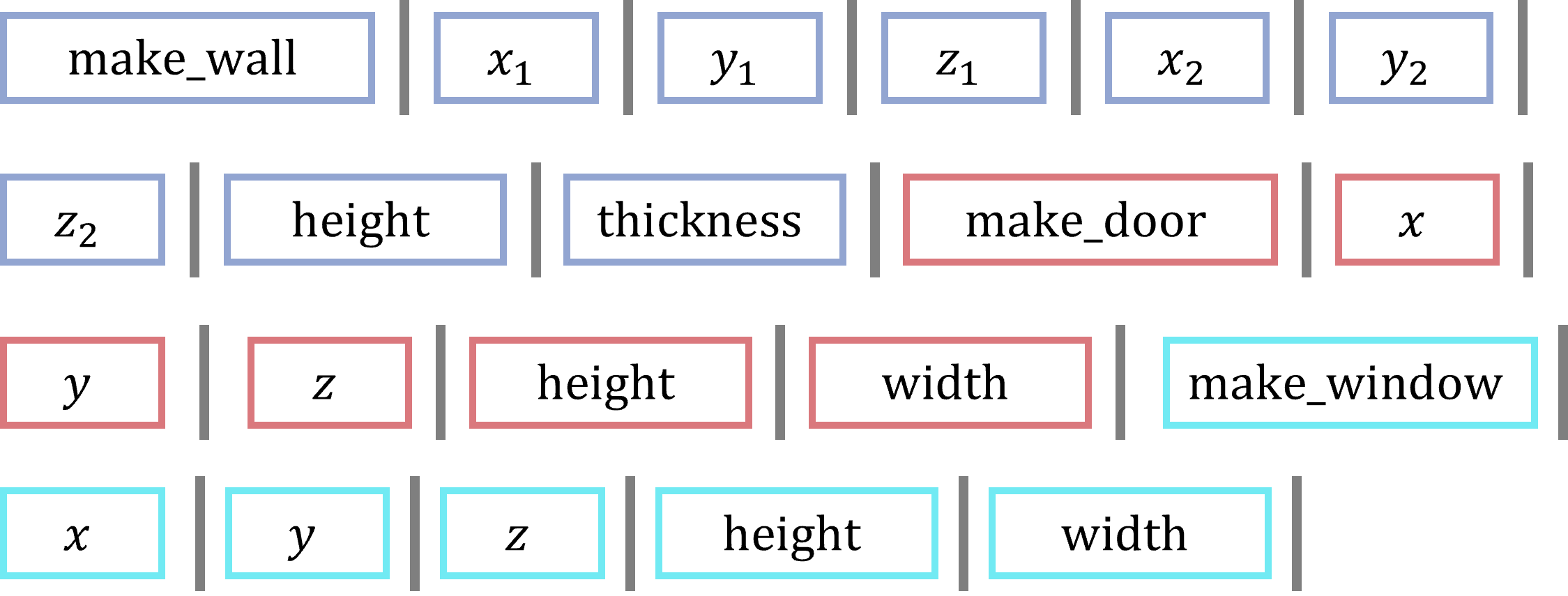}
    \caption{Inference process by SceneScript~\cite{avetisyan2024SceneScript} ($\textbf{21}$ iterations)}
    \label{fig:mtp_concept_b}
  \end{subfigure}\hfill
  \begin{subfigure}[t]{0.385\linewidth}
    \centering
    \includegraphics[width=\linewidth]{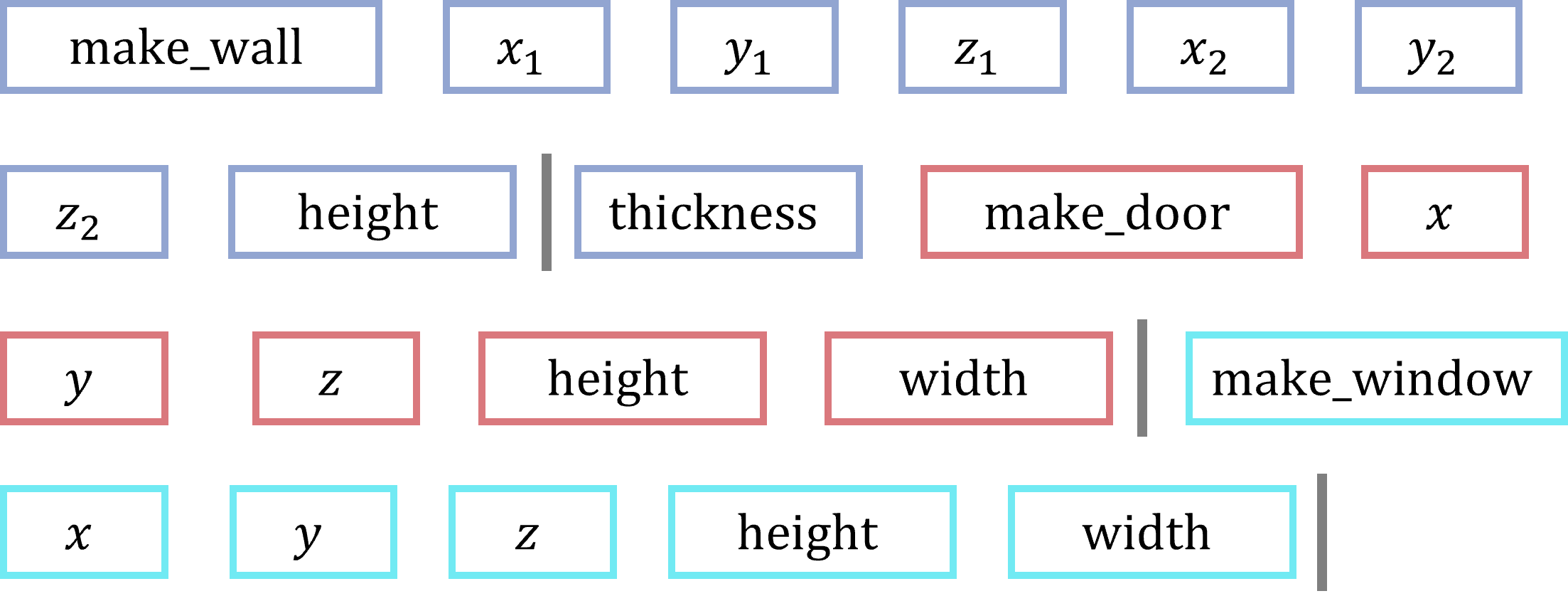}
    \caption{Inference process by Fast SceneScript ($\textbf{3}$ iterations)}
    \label{fig:mtp_concept_c}
  \end{subfigure}

   \caption{Comparison between our proposed Fast SceneScript and SceneScript~\cite{avetisyan2024SceneScript}. Gray vertical line in (b) and (c) indicates the separation between different forward passes. To generate the same sequence, SceneScript~\cite{avetisyan2024SceneScript} requires $21$ decoder iterations, whereas our method completes this in just $3$ iterations (producing $8$, $7$, and $6$ tokens per iteration), delivering significantly higher efficiency.}
   \label{fig:mtp_concept}
\end{figure*}

To enable compatibility with language models, structured languages are designed to represent 3D scenes by encoding geometric and semantic properties as sequences of tokens following a predefined schema. For example, SceneScript~\cite{avetisyan2024SceneScript, DBLP:journals/corr/abs-2503-11806} represents a wall as a list of tokens, \ie, $[make\_wall, x_1, y_1, z_1, x_2, y_2, z_2, height, thickness]$, in which $[x_1, y_1, z_1]$ and $[x_2, y_2, z_2]$ are the coordinates of two corners. Following language models~\cite{vaswani2017attention, DBLP:journals/corr/abs-2302-13971, DBLP:journals/corr/abs-2412-19437}, next-token prediction (NTP) is used to predict tokens one-by-one, leading to a versatile and flexible framework. However, predicting one token per decoder inference in NTP causes high latency and becomes particularly inefficient when the sequence length increases.

Unlike natural languages, the structured language for visual perception is more deterministic and weakly coupled, making it more predictable. This property enables the feasibility of predicting multiple tokens per decoder pass, \ie, multi-token prediction (MTP)~\cite{gloeckle2024better}. MTP can reduce the number of autoregressive iterations and thus improve the inference efficiency significantly. However, since predicting multiple tokens simultaneously is challenging, MTP models suffer from accuracy degradation, making them less competitive than NTP models, as shown in~\cref{fig:teaser}. Besides, the MTP module requires additional token prediction heads, which introduces a large number of parameters.

Motivated by these observations, we propose Fast SceneScript, a novel framework for 3D scene understanding that achieves a $5.09\times$ speed-up for layout estimation and a $5.14\times$ speed-up for object detection while preserving accuracy compared to SceneScript~\cite{avetisyan2024SceneScript}. To enable this speed-up, we employ multi-token prediction (see \cref{fig:mtp_concept}). To address accuracy degradation from unreliable token predictions, we adapt self-speculative decoding (SSD)~\cite{DBLP:conf/acl/Zhang00S0CM24,stern2018blockwise} and propose confidence-guided decoding (CGD). In particular, to accommodate structured language models, we augment SSD by applying a distance metric for numerical tokens to improve acceleration. CGD proposes a new scoring method to better assess token reliability on the fly. Finally, to address the parameter overhead introduced by MTP heads, we design a parameter-efficient mechanism, which enables effective feature extraction for additional heads.

To the best of our knowledge, this is the first work to introduce multi-token prediction to language-based perception models, leading to accurate and efficient inference. Our key contributions can be summarized as follows:
\begin{itemize}
    \item We propose a novel structured language model employing multi-token prediction for efficient inference.
    \item We investigate the decoding strategies with filtering mechanisms for structured language models to achieve accurate and reliable inference. 
    \item We design a parameter-efficient mechanism in MTP models, which reduces the number of parameters by $\sim43\%$ while maintaining accuracy.
    \item The experimental results on ASE~\cite{avetisyan2024SceneScript}, Structured3D~\cite{zheng2020structured3d}, and SceneCAD~\cite{avetisyan2020scenecad} datasets demonstrate that, compared to SceneScript~\cite{avetisyan2024SceneScript}, Fast SceneScript can predict around 9 tokens per step on average without compromising accuracy, while increasing the number of parameters by only $\sim7.5\%$. Furthermore, Fast SceneScript achieves $5.09\times$ and $5.14\times$ faster decoding than SceneScript~\cite{avetisyan2024SceneScript} in layout estimation and object detection, respectively, setting a new benchmark for language-based perception models.
\end{itemize}
\section{Related Work}
\label{sec:relatedwork}

\subsection{Language-based Perception Generalist Model}
Due to the versatility and flexibility of the language models~\cite{chenpix2seq,chen2022unified,avetisyan2024SceneScript,DBLP:journals/corr/abs-2411-09595,DBLP:journals/corr/abs-2508-14879,DBLP:conf/cvpr/HuangZLGZ025,xue2022point2seq,DBLP:conf/cvpr/YinRYDH25,DBLP:journals/corr/abs-2507-18300,DBLP:conf/eccv/WangTJSNLSW24}, there is a trend to use language as a unified interface for different perception tasks. In 2D perception, there are various tasks, such as object detection~\cite{DBLP:conf/nips/RenHGS15, DBLP:conf/eccv/CarionMSUKZ20}, instance segmentation~\cite{DBLP:conf/iccv/HeGDG17,DBLP:conf/cvpr/0004GGH18}, keypoint detection~\cite{DBLP:conf/iccv/HeGDG17,DBLP:conf/cvpr/0004GGH18}, and image captioning~\cite{vaswani2017attention,DBLP:conf/cvpr/PanYLM20}. Pix2Seq \cite{chenpix2seq,chen2022unified} formulates the output of each task as a sequence of discrete tokens with a unified interface and shows that a single neural network can be applied to solve all these tasks. VisionLLM~\cite{DBLP:conf/nips/WangCCWZZLLZQD23} and Rex-Omni~\cite{DBLP:journals/corr/abs-2510-12798} propose multimodal large language models that define object detection~\cite{DBLP:conf/nips/RenHGS15} and other visual perception tasks~\cite{DBLP:conf/cvpr/SinghPT0GH21,DBLP:conf/lrec/LiCHWZL20,DBLP:conf/cvpr/LuoZZ0T22} as a simple next-token prediction problem.

\begin{figure*}[htbp]
  \centering
  \includegraphics[width=0.9\linewidth]{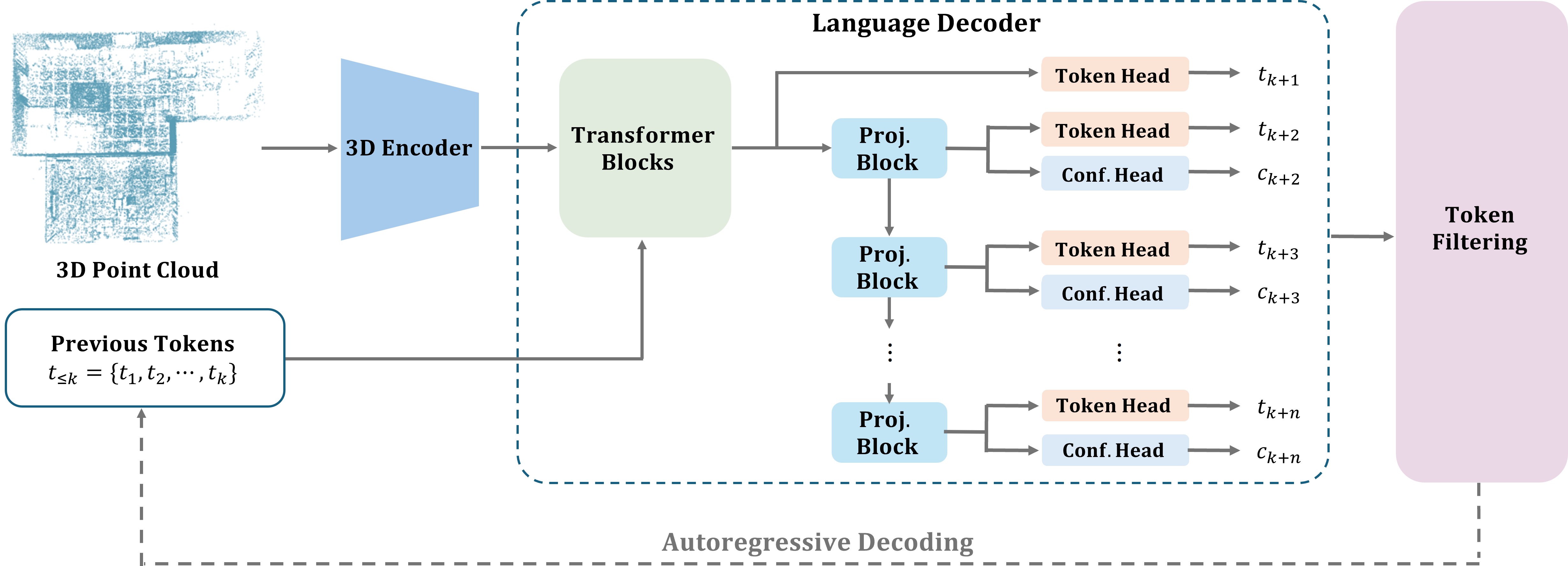}

   \caption{Framework of our proposed Fast SceneScript. The 3D point cloud is initially processed by the 3D encoder. Subsequently, the extracted 3D features, together with a sequence of preceding tokens, \ie, $\{t_{1}, t_{2}, \dots, t_{k}\}$, are fed into the language decoder to generate the next $n$ tokens $\{t_{k+1}, t_{k+2}, \dots, t_{k+n}\}$ and $(n-1)$ confidences $\{c_{k+2}, ..., c_{k+n}\}$. Projection block, token head, and confidence head are shared across all $n$ tokens. To improve prediction reliability, token filtering strategy is applied to remove unreliable tokens.}
   \label{fig:framework}
\end{figure*}

In 3D perception, SceneScript~\cite{avetisyan2024SceneScript, DBLP:journals/corr/abs-2503-11806}, an autoregressive structured language model, considers the 3D perception as a task of sequence generation, in which structured languages are designed to represent the output of various tasks. SceneScript~\cite{avetisyan2024SceneScript, DBLP:journals/corr/abs-2503-11806} can perform layout estimation~\cite{yue2023connecting}, object detection~\cite{DBLP:conf/cvpr/Brazil0SR0G23,DBLP:conf/iccv/MisraGJ21,DBLP:conf/wacv/RukhovichV022}, and coarse 3D object part reconstruction~\cite{DBLP:conf/cvpr/TulsianiSGEM17,DBLP:journals/tog/YangC21} with a single unified architecture and interface. SpatialLM~\cite{mao2025spatiallm} is a large language model which processes 3D point cloud data and generates structured 3D scene understanding outputs. It supports both layout estimation~\cite{yue2023connecting} and object detection~\cite{DBLP:conf/iclr/0001GYLLW00G24} tasks.

These existing methods follow the next-token prediction paradigm, \ie, predicting one token in a single inference, which leads to slow inference speed. In contrast, our Fast SceneScript significantly improves the efficiency of SceneScript~\cite{avetisyan2024SceneScript}, a representative language-based perception model, without harming its versatility and flexibility.

\subsection{Multi-token Prediction and Token Filtering}
In large language model (LLM) community, multi-token prediction (MTP)~\cite{gloeckle2024better,stern2018blockwise}, \ie, generating multiple tokens in one forward inference, is proposed to accelerate the inference speed. For example, DeepSeek-V3~\cite{DBLP:journals/corr/abs-2412-19437}, Medusa~\cite{cai2024medusa}, and VocalNet~\cite{DBLP:journals/corr/abs-2504-04060} integrate additional heads to predict multiple tokens. However, MTP models often generate inaccurate tokens, which can significantly decrease accuracy. Therefore, decoding algorithms to identify and filter out unreliable tokens are applied. Self-speculative decoding (SSD)~\cite{DBLP:conf/acl/Zhang00S0CM24,DBLP:conf/icml/LeviathanKM23,stern2018blockwise} first generates multiple candidate tokens and then compare them with the NTP results in the next inference iteration. The longest consistent prefix is accepted. Differently, scoring-based methods~\cite{DBLP:conf/naacl/TuliLHJSJ24,DBLP:journals/corr/abs-2509-24007} predict multiple candidate tokens and their scores in the same inference iteration. The longest reliable prefix is determined by applying a threshold to the scores, which achieves on-the-fly decoding. Specifically, the cumulative product of softmax probability is used as the score metric in~\cite{DBLP:conf/naacl/TuliLHJSJ24,DBLP:journals/corr/abs-2509-24007}.

In this work, we investigate both SSD and scoring-based approaches to enhance accuracy. To accommodate structured language models, we augment SSD by applying a distance metric for numerical tokens to further accelerate inference. We advance scoring-based methods~\cite{DBLP:conf/naacl/TuliLHJSJ24,DBLP:journals/corr/abs-2509-24007} by proposing a new scoring to better assess token reliability. Additionally, we design a parameter-efficient mechanism that substantially reduces the parameter overhead of MTP.

\section{Method}
\label{sec:method}

This work aims to accurately and efficiently predict structured language tokens for 3D scene understanding from 3D point clouds. To this end, we employ three key components: (1) multi-token prediction to decode tokens in parallel, achieving faster inference, (2) unreliable token filtering to improve accuracy, (3) a parameter-efficient mechanism to reduce the number of parameters in MTP heads~\cite{gloeckle2024better}.

The proposed framework is illustrated in~\cref{fig:framework}. First, the 3D point cloud is encoded by a sparse 3D ResNet~\cite{tang2020searching,DBLP:conf/mlsys/TangLL0022} to generate 3D features. Then, conditioned on 3D features and previous tokens $\{t_1, t_2, \dots, t_k\}$, our Fast SceneScript generates reliable future tokens, through token prediction and token filtering stages. During token prediction, the language decoder generates $n$ future tokens, \ie, $\{t_{k+1}, t_{k+2}, \dots, t_{k+n}\}$. The language decoder, which is composed of self-attention and cross-attention layers, utilizes Transformer blocks~\cite{vaswani2017attention} to yield language hidden state $f_{k+1}$. Afterwards, the shared projection block is applied on $f_{k+1}$ to generate distinct hidden state $f_{k+i}$ for additional future tokens, $i \in [2, n]$. Next, these token features go through the shared token head and confidence head to predict future tokens $\{t_{k+1}, t_{k+2}, \dots, t_{k+n}\}$ and confidences $\{c_{k+2}, \dots, c_{k+n}\}$. Finally, an additional token filtering stage is adopted to discard unreliable tokens.

\subsection{Multi-token Prediction}
{\bf Next-token prediction} For a 3D scene represented as structured language of $N$ tokens, given the previous tokens $t_{\leq k}=\{t_1, t_2, \dots, t_k\}, k \in [1, N]$, the existing language model~\cite{avetisyan2024SceneScript} sequentially predicts the probability $p(t_{k+1}|t_{\leq k})$ for the next token. During training, cross-entropy loss is adopted, \ie,
\begin{equation}
    \mathcal{L}_{NTP} = -\sum_{k} \text{log }{p(t_{k+1}|t_{\leq k})}
\end{equation}
During inference, a sampling strategy is applied to $p(t_{k+1}|t_{\leq k})$ to generate a discrete token $t_{k+1}$, such as greedy decoding and nucleus sampling.

{\bf Multi-token prediction} 
To improve the efficiency of language models, multi-token prediction~\cite{gloeckle2024better} generates probabilities $\{p(t_{k+i}|t_{\leq k})\}_{i=1}^n$ for $n$ future tokens at once, predicted by $n$ token heads. The training objective of MTP is expressed as follows,
\begin{equation}
    \mathcal{L}_{MTP} = -\sum_{k} \sum_{i} \lambda_h^{i-1}  \text{log }{ p(t_{k+i}|t_{\leq k})}
\end{equation}
where $\lambda_h^{i-1}$ denotes the loss weight for token head $i$, following the approaches~\cite{cai2024medusa,DBLP:journals/corr/abs-2504-04060}. As $i$ increases from 1 to $n$, the output becomes more uncertain and incurs a higher loss. To mitigate the negative impact of subsequent heads during training, $\lambda_h$ is designed as a decaying factor. 

During inference, MTP requires only $\lceil N / n \rceil$ forward passes of the language decoder to generate a sequence of $N$ tokens, compared to $N$ passes for NTP, thereby significantly accelerating inference.

\begin{figure}[t]
  \centering
    \begin{subfigure}[t]{0.75\linewidth}
    \centering
    \includegraphics[width=\linewidth]{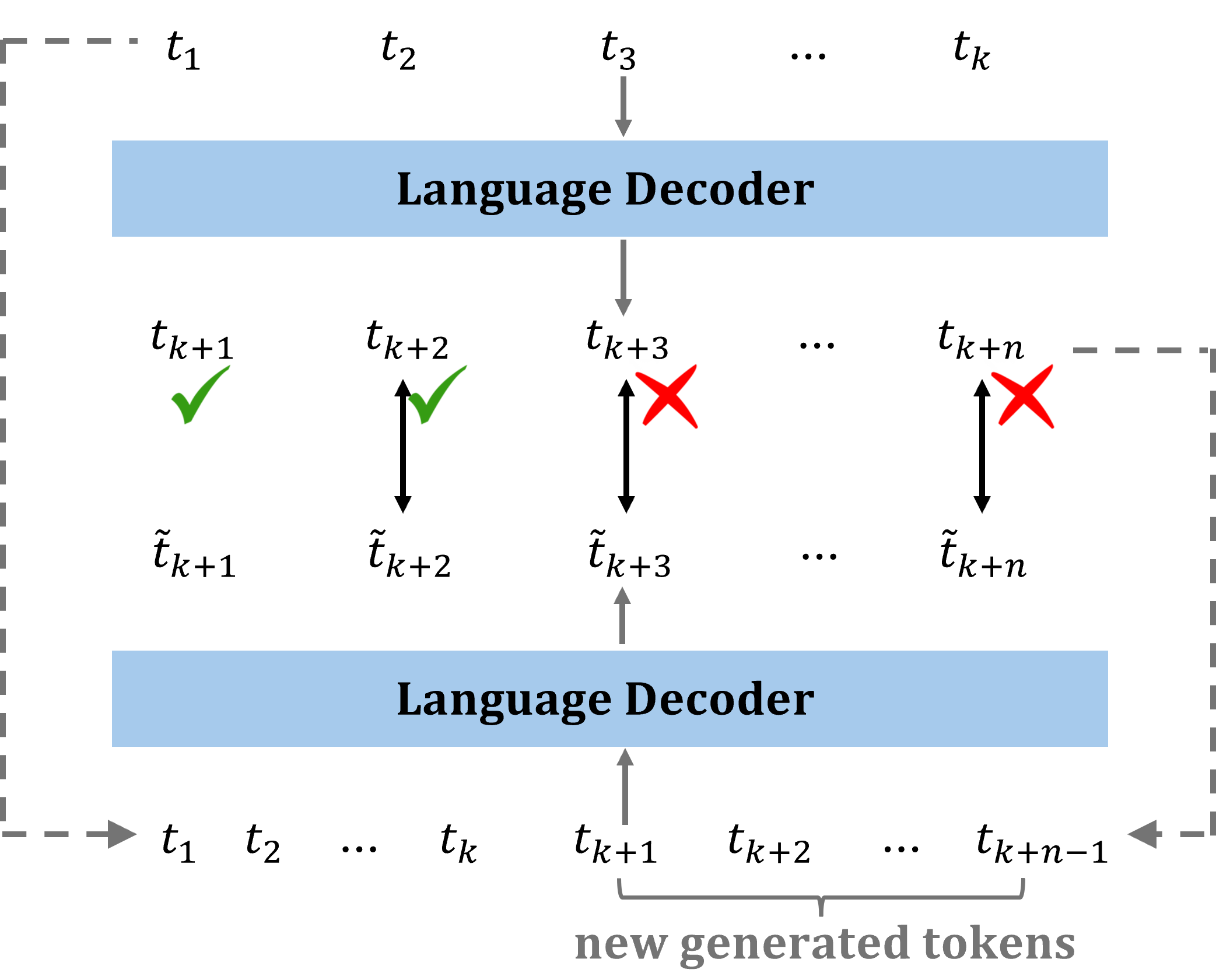}
    \caption{Self-Speculative Decoding (SSD)~\cite{DBLP:conf/acl/Zhang00S0CM24,stern2018blockwise}}
    \label{fig:token_filtering_a}
  \end{subfigure}

  \vspace{0.5em}

  \begin{subfigure}[t]{0.62\linewidth}
    \centering
    \includegraphics[width=\linewidth]{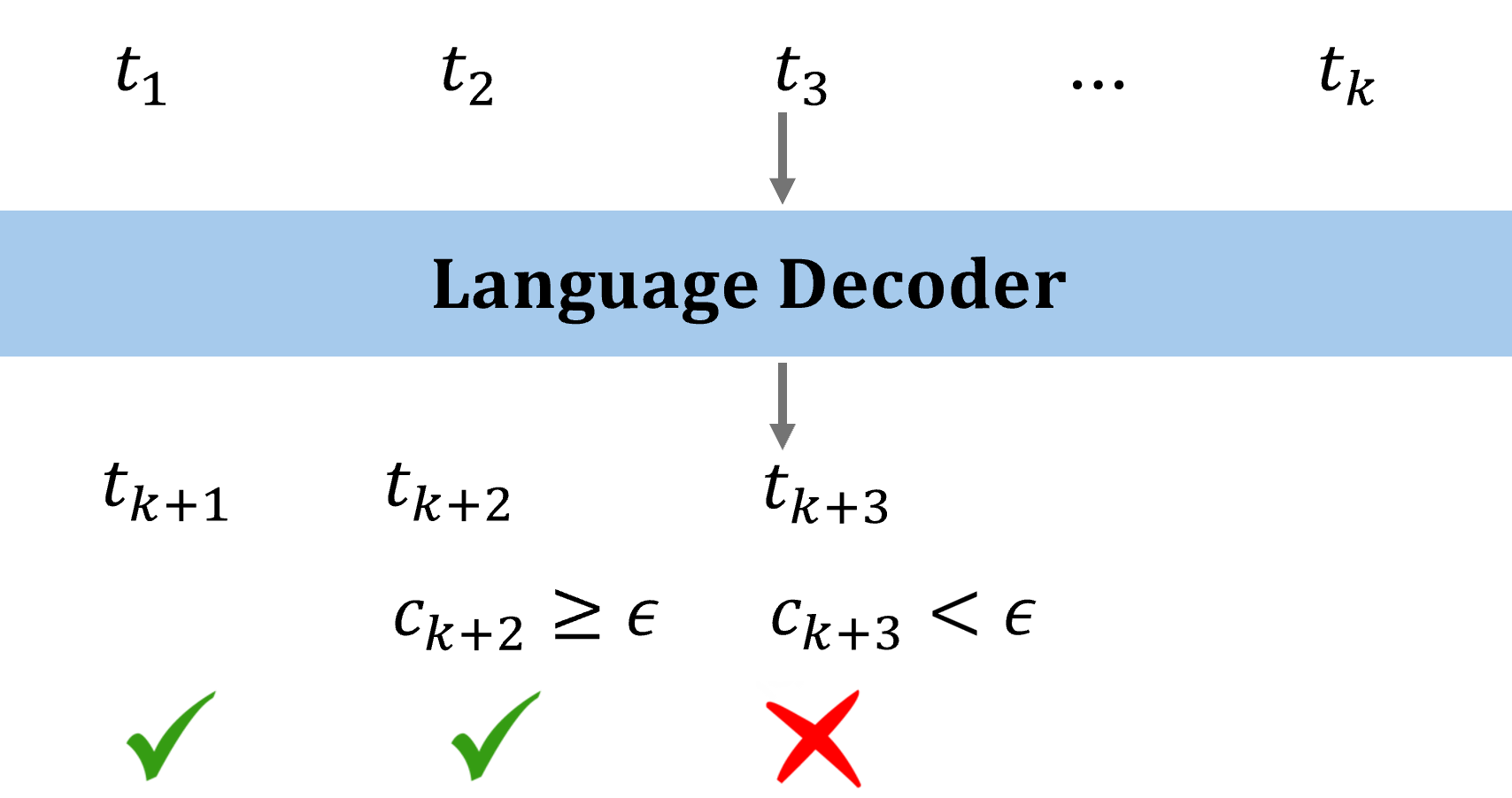}
    \caption{Confidence-Guided Decoding (CGD)}
    \label{fig:token_filtering_b}
  \end{subfigure}
   \caption{Token filtering strategies. This figure illustrates an example where the first two generated tokens are accepted. (a): The language decoder drafts next $n$ tokens $\{t_{k+1}, t_{k+2}, ..., t_{k+n}\}$ conditioned on preceding tokens $t_{\leq k}$. Then, $t_{\leq k+n-1}$ are fed to the language decoder again for next-token prediction, generating $\{\widetilde{t}_{k+i}\}_{i=1}^n$ using the first token head. Finally, consistency checking is applied between $\widetilde{t}_{k+i}$ and $t_{k+i}$, after which only tokens before the first unreliable token are retained. (b): The language decoder predicts both tokens and their confidences. Decoding terminates upon encountering the first token deemed unreliable.}
   \label{fig:token_filtering}
\end{figure}

\subsection{Unreliable Token Filtering}
Although the MTP model can achieve a significant latency speedup, predicting multiple tokens in one inference is challenging.
Thus, the accuracy of the MTP approach can be significantly worse than the NTP model.
To enable both fast inference and accurate prediction, this work employs token filtering, which further measures the reliability of MTP outputs and filters out unreliable tokens (see~\cref{fig:token_filtering}).

We explore two types of token filtering strategies, 
\ie, self-speculative decoding (SSD)~\cite{DBLP:conf/acl/Zhang00S0CM24,DBLP:conf/icml/LeviathanKM23,stern2018blockwise} and our proposed confidence-guided decoding (CGD). SSD initially drafts $n$ candidate tokens in the first iteration. In the following iteration, these tokens are verified, and their consistency across the two iterations is evaluated. Only consistent tokens are considered reliable and accepted. This process enables SSD to validate predictions with a one-step delay. Differently, CGD proposes to predict the tokens and their confidences simultaneously, in which the confidence corresponds to the token reliability. In this manner, CGD can determine where to stop on the fly, reducing latency and unnecessary computation for unreliable tokens.

{\bf Self-speculative decoding} During inference, SSD first drafts $n$ future tokens $\{t_{k+1}, t_{k+2}, ..., t_{k+n}\}$ at once by $n$ MTP token heads, based on the previous context $t_{\leq k}=\{t_1, t_2, ..., t_k\}$. To verify their accuracy, the model extends the sequence to $\{t_1, t_2, ..., t_{k+n-1}\}$ and feeds it back into the network for next-token prediction, generating $\{\widetilde{t}_{k+2}, \widetilde{t}_{k+3}, ..., \widetilde{t}_{k+n}\}$ with the first token head. Each predicted token $\widetilde{t}_{k+i}$ is conditioned on the preceding context $\{t_1, t_2, \dots, t_{k+i-1}\}, i \in [2, n]$. Afterwards, the predicted tokens $\{t_{k+2}, ..., t_{k+n}\}$ from the first iteration are compared against their corresponding counterparts $\{\widetilde{t}_{k+2}, \widetilde{t}_{k+3}, ..., \widetilde{t}_{k+n}\}$ from the second iteration to assess the alignment. Since token $t_{k+1}$ predicted from the first head deemed reliable, no additional verification is needed. To accommodate the structured language model, We employ a distance metric for numerical tokens, such as $x_1$ and $height$, to accommodate the structured language model. The motivation is that a small margin for numerical tokens is acceptable, which also improves the number of accepted tokens per decoder inference. Numerical tokens that satisfy the condition in~\cref{eq:distance_metric} are deemed reliable; otherwise, they are marked as unreliable.
\begin{equation}
\begin{aligned}
    |t_{k+i} - \widetilde{t}_{k+i}|\leq \tau
\end{aligned}
\label{eq:distance_metric}
\end{equation}
Here, $\tau$ is a positive hyperparameter. For non-numerical tokens such as $part$, $make\_wall$, and $stop$, reliability is evaluated by checking for exact equality.

{\bf Confidence-guided decoding}
We propose a novel decoding strategy, \ie, confidence-guided decoding (CGD). Its objective is to measure the reliability of tokens while eliminating verification delays, thereby achieving verification on the fly and removing the computational costs for unreliable tokens. Since the first head’s prediction is typically more reliable, confidence heads estimate the consistency of additional heads’ predictions with that of the first head.

\emph{During training}, the ground truth $\hat{c}_{k+i}$ for confidence head $i$ is generated by measuring the alignment between output $\widetilde{t}_{k+i}$ generated by the first head and output $t_{k+i}$ generated by head $i$, $i \in [2, n]$. If $t_{k+i}$ satisfies~\cref{eq:distance_metric}, $\hat{c}_{k+i}=1$, otherwise $\hat{c}_{k+i}=0$. With $c_{k+i}$ denoting the predicted confidence for $t_{k+i}$, binary cross-entropy loss is applied as confidence loss,

\begin{equation}
\begin{aligned}
    \mathcal{L}_c = -\sum_{i, k} \lambda_h^{i-1} \Big( 
    \hat{c}_{k+i} \log c_{k+i} 
    + (1 - \hat{c}_{k+i}) \log (1 - c_{k+i}) 
    \Big),
\end{aligned}
\end{equation}

Overall, the training objective of our method is, 
\begin{equation}
    \mathcal{L} = \mathcal{L}_{MTP} + \lambda_c \mathcal{L}_c 
\end{equation}
where  $\lambda_c$ denotes loss weight for confidence loss. Additional details can be found in the supplementary material.

\emph{During inference}, our CGD predicts $n$ tokens $\{t_{k+i} \}_{i=1}^{n}$ and $(n-1)$ confidences $\{c_{k+2}, ..., c_{k+n}\}$. Then, we verify the reliability of the predicted tokens by checking their confidence scores. The tokens satisfying $c_{k+i} < \epsilon$ are considered as unreliable tokens, and we accept the longest reliable prefix. In this way, we predict, verify, and accept tokens in a single inference iteration, enabling on-the-fly decoding.

\begin{table*}[t]
  \caption{Quantitative comparisons for layout estimation on ASE dataset~\cite{avetisyan2024SceneScript}. $*$ denotes the results evaluated with the author-released model, whereas all other results are obtained from models trained in this paper. $n$ denotes the number of MTP heads. $\alpha_{val}$ and $\alpha_{test}$ refer to the average number of tokens accepted per decoder inference on \emph{val} and \emph{test} sets.
  }
  \label{tab:results_aria}
  \small
  \centering
  \setlength{\tabcolsep}{3pt}
  \begin{tabular}{llccc|c|cccc|c|cccc}
    \toprule

   & \multirow{2}{*}{Method} & \multirow{2}{*}{$n$} & \multirow{2}{*}{Param $\downarrow$} & \multirow{2}{*}{Latency $\downarrow$} & \multirow{2}{*}{$\alpha_{val}$ $\uparrow$} & \multicolumn{4}{c|}{F1-Score of \emph{val} set $\uparrow$} & \multirow{2}{*}{$\alpha_{test}$ $\uparrow$} & \multicolumn{4}{c}{F1-Score of \emph{test} set $\uparrow$} \\
    & & & & & & wall & window & door & mean & & wall  & window  & door & mean\\
     \midrule
   a& SceneScript $^*$~\cite{avetisyan2024SceneScript} & 1 & 14.00 M & 387 ms &1 & 0.890 & 0.865 & 0.918 & 0.891 & 1 &  0.896 & 0.865  & 0.928 & 0.896\\
    b&SceneScript~\cite{avetisyan2024SceneScript} & 1 & 14.00 M & 382 ms &1 & 0.918 & 0.880 & 0.940 & 0.913 & 1 & 0.921 & 0.881  & 0.942 & 0.915 \\
    \midrule
   c& SceneScript~\cite{avetisyan2024SceneScript} + MTP~\cite{gloeckle2024better} & 4 & 18.14 M & 109 ms & 4 & 0.891 & 0.852 & 0.913 & 0.885 &4 & 0.898  & 0.855  & 0.913 & 0.889\\

    \midrule
   d& SceneScript~\cite{avetisyan2024SceneScript} + MTP~\cite{gloeckle2024better} & 6 & 20.91 M  & 76 ms & 6 & 0.835 & 0.813 & 0.880 & 0.843 &6 & 0.838  & 0.813 & 0.889 & 0.847 \\

    \midrule
   e& SceneScript~\cite{avetisyan2024SceneScript} + MTP~\cite{gloeckle2024better} & 8 & 23.67 M  & 62 ms & 8& 0.831 & 0.804 & 0.885 & 0.840 &  8  & 0.836  & 0.804 & 0.886 & 0.842\\

   f& Fast SceneScript (SSD) & 8  & 15.05 M & 81 ms & 7.46 & 0.914 &0.882 & 0.939 & 0.912 & 7.45  & 0.919  & 0.882 & 0.939 & 0.913\\

  g&  Fast SceneScript (CGD) & 8 & 16.10 M & 92 ms &6.29 & 0.912 & 0.883 & 0.938 &0.911 & 6.30 & 0.918 & 0.883  & 0.938 & 0.913 \\
    \midrule
  
  h& SceneScript~\cite{avetisyan2024SceneScript} + MTP~\cite{gloeckle2024better} & 10 & 26.43 M  & 54 ms &10 & 0.805 & 0.776 & 0.863 & 0.815 & 10  & 0.808 & 0.774 & 0.861 & 0.814\\
   i& Fast SceneScript (SSD) & 10 & 15.05 M &75 ms& 8.97 & 0.910 & 0.879 & 0.937 & 0.909&  8.99 & 0.915 & 0.880 &  0.940 & 0.912\\
   j& Fast SceneScript (CGD) & 10 & 16.10 M & 89 ms & 7.27& 0.909 & 0.880 & 0.936 & 0.908& 7.27 & 0.912  & 0.879  & 0.938 & 0.910 \\
    \bottomrule
  \end{tabular}
\end{table*}

\subsection{Parameter-Efficient Mechanism}

MTP~\cite{gloeckle2024better} predicts $n$ tokens in parallel, which introduces $(n - 1)$ extra token heads for future tokens. To improve parameter efficiency, we introduce a head-sharing parameter-efficient mechanism (illustrated in~\cref{fig:framework}), where $n$ heads share the same parameters. This approach leverages the language model’s strength and aligns with autoregressive decoding: hidden states remain (1) context-dependent—each encodes unique position-specific context despite sharing the token head—and (2) embedded in a shared semantic space expressive enough for diverse predictions.

To generate distinct hidden states for $n-1$ additional heads, we introduce a lightweight, parameter-efficient feature projection block (see~\cref{fig:framework}). It consists of two feed-forward blocks, each with 2 linear layers, 1 ReLU activation, and 1 layer normalization—similar to Transformer FFNs~\cite{vaswani2017attention}. Shared across all heads, this block refines token representations with minimal overhead while preserving the flow of contextual features.

\section{Experiments}
\label{sec:exps}

\begin{table*}
  \caption{Quantitative comparisons for layout estimation on Structured3D \emph{test} set \cite{zheng2020structured3d}.}
  \label{tab:results_stru3d}
  \small
  \centering
  \setlength{\tabcolsep}{7pt}
  \begin{tabular}{clcccc|cccc}
    \toprule

    &\multirow{2}{*}{Method} & \multirow{2}{*}{$n$} & \multirow{2}{*}{Param $\downarrow$} & \multirow{2}{*}{Latency $\downarrow$} & \multirow{2}{*}{$\alpha_{test}$ $\uparrow$} & \multicolumn{4}{c}{F1-Score $\uparrow$}\\
    &  &  &  &  & & wall & window & door & mean \\

     \midrule
  a & RoomFormer~\cite{yue2023connecting} & - &  17.17 M & 54 ms &- & 0.758 & 0.608 & 0.739 & 0.702 \\
    \midrule
   b& SceneScript~\cite{avetisyan2024SceneScript} & 1 & 14.00 M & 1176 ms &1 & 0.801 & 0.656 & 0.864 & 0.774 \\
    \midrule
   c& SceneScript~\cite{avetisyan2024SceneScript} + MTP~\cite{gloeckle2024better} & 4 & 18.14 M & 315 ms  & 4 & 0.753 & 0.676 & 0.864 & 0.764 \\

    \midrule
   d& SceneScript~\cite{avetisyan2024SceneScript} + MTP~\cite{gloeckle2024better} & 6 & 20.91 M & 217 ms & 6 & 0.751 & 0.665 & 0.860 & 0.759 \\

    \midrule
   e& SceneScript~\cite{avetisyan2024SceneScript} + MTP~\cite{gloeckle2024better} & 8 & 23.67 M & 171 ms & 8& 0.740 & 0.662 & 0.853 & 0.752 \\
  f & Fast SceneScript (SSD) & 8 & 15.05 M &230 ms& 7.40 & 0.800 &0.692 & 0.881 & 0.791 \\

   g& Fast SceneScript (CGD) & 8 & 16.10 M & 269 ms& 6.15 &0.803  & 0.704 & 0.879 & 0.795\\
    \midrule
    h& SceneScript~\cite{avetisyan2024SceneScript} + MTP~\cite{gloeckle2024better} & 10 & 26.43 M & 154 ms &10 & 0.711 & 0.638 & 0.821 & 0.723 \\
  i & Fast SceneScript (SSD) & 10 & 15.05 M & 211 ms & 8.86 & 0.803 & 0.691 & 0.877 & 0.790 \\
   j& Fast SceneScript (CGD) & 10 & 16.10 M & 257 ms & 7.15 & 0.797 & 0.690 & 0.882 & 0.790 \\
    \bottomrule
  \end{tabular}
\end{table*}

\begin{table*}[h]
  \caption{Quantitative comparisons for layout estimation on SceneCAD \emph{val} set~\cite{avetisyan2020scenecad}. 
  }
  \label{tab:sota_scenecad}
  \small
  \centering
  \begin{tabular}{llc|ccc|cccc}
    \toprule

&\multirow{2}{*}{Method}  & \multirow{2}{*}{$n$}& \multicolumn{3}{c|}{Layout Estimation} &  \multicolumn{4}{c}{Object Detection} \\
 & & &  Latency $\downarrow$ & $\alpha$ $\uparrow$ & F1-Score $\uparrow$ & Latency $\downarrow$ & $\alpha$ $\uparrow$ & F1 @.25 $\uparrow$ & F1 @.50 $\uparrow$\\
     \midrule
 a&SceneScript~\cite{avetisyan2024SceneScript} & 1 & 108 ms &1 & 0.556 & 104 ms &  1 & 0.670 & 0.461 \\
    \midrule

 b&SceneScript~\cite{avetisyan2024SceneScript} + MTP~\cite{gloeckle2024better} & 8 & 23 ms & 8& 0.460 & 20 ms & 8 & 0.688 & 0.508 \\
c &  Fast SceneScript (SSD)  & 8 & 42 ms& 5.07 & 0.556 &33 ms & 5.72 & 0.703  & 0.544 \\

d &Fast SceneScript (CGD)  & 8 & 46 ms& 4.87 &0.552  & 31 ms & 5.68 & 0.710  & 0.529 \\
    \bottomrule
  \end{tabular}
  
\end{table*}

\begin{table*}[htbp]
  \caption{Ablation experiments for layout estimation on the ASE \emph{val} set~\cite{avetisyan2024SceneScript}. 
  }
  \label{tab:ab_aria}
  \small
  \centering
  \setlength{\tabcolsep}{5pt}
  \begin{tabular}{clccc|ccccc}
    \toprule

   &\multirow{2}{*}{Method} & \multirow{2}{*}{$n$} & \multirow{2}{*}{Param $\downarrow$} & \multirow{2}{*}{Latency $\downarrow$} & \multirow{2}{*}{$\alpha_{val}$ $\uparrow$} & \multicolumn{4}{c}{F1-Score $\uparrow$}\\
    &  &  &  &  & & wall & window & door & mean \\
     \midrule
   a& SceneScript~\cite{avetisyan2024SceneScript} & 1 & 14.00 M &  375 ms &1 & 0.918 & 0.880  & 0.940 &0.913 \\

   b& SceneScript~\cite{avetisyan2024SceneScript} + MTP~\cite{gloeckle2024better} & 8 & 23.67 M&  61 ms & 8& 0.831  & 0.804 & 0.885 & 0.840\\ \hline
   
   & \textit{SSD-based methods} &  &   &  &  &&  &  &  \\
   c& Fast SceneScript (SSD, $\tau=0$) & 8 & 15.05 M & 90 ms & 6.53  & 0.913  & 0.882 & 0.939 & 0.911 \\ 
   d& Fast SceneScript (SSD, $\tau=2$) & 8 & 15.05 M & 80 ms & 7.46  & 0.914  & 0.882 & 0.939 & 0.912 \\ 
   e& Fast SceneScript (SSD, $\tau=5$) & 8 & 15.05 M & 80 ms & 7.52 & 0.913  & 0.881 & 0.938 & 0.911 \\
   f& Fast SceneScript (SSD, $\tau=2$, w/o sharing) & 8 &  23.67 M& 75 ms & 7.34 & 0.909 & 0.880 & 0.935 & 0.908 \\ \hline
   & \textit{Scoring-based methods} &  &   &  &  &  &  &  \\
   g& Fast SceneScript (ProductThre~\cite{DBLP:conf/naacl/TuliLHJSJ24, DBLP:journals/corr/abs-2509-24007}) & 8 & 15.05 M & 112 ms & 4.67 & 0.908 & 0.881 & 0.938 & 0.909 \\
   h& Fast SceneScript (SoftmaxThre~\cite{DBLP:journals/corr/abs-2502-09992}) & 8 & 15.05 M & 116 ms & 4.53 & 0.908 & 0.880 & 0.937 & 0.908 \\
   i& Fast SceneScript (CGD) & 8 & 16.10 M &  91 ms & 6.29 & 0.912 & 0.883  & 0.938 & 0.911 \\
   j& Fast SceneScript (CGD, w/o sharing) & 8 & 31.03 M& 89 ms& 6.25 & 0.913 & 0.880 & 0.936 & 0.910 \\
    \bottomrule
  \end{tabular}
\end{table*}

\begin{figure*}[htbp]
  \centering
  \includegraphics[width=0.9\linewidth]{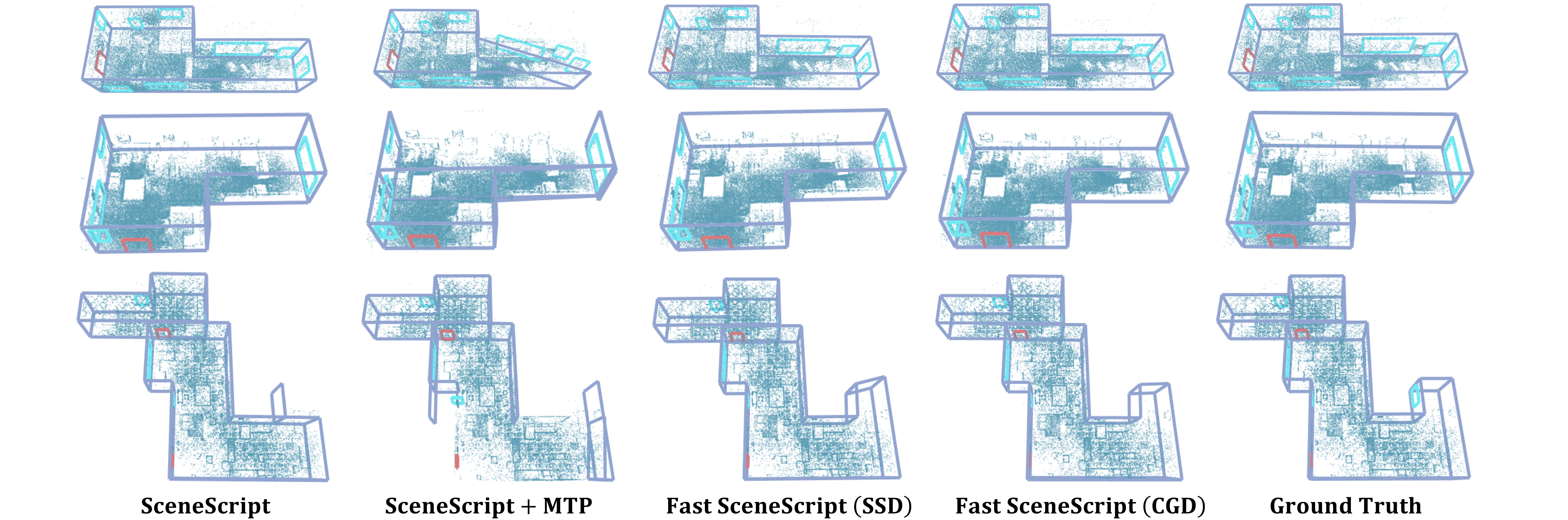}

   \caption{Qualitative comparisons on ASE \emph{test} set~\cite{avetisyan2024SceneScript}. The number of MTP heads $n$ is set to 8. Compared to SceneScript~\cite{avetisyan2024SceneScript} + MTP~\cite{gloeckle2024better}, our Fast SceneScript can predict more complete and accurate scene layouts.}
   \label{fig:vis_aria}
\end{figure*}


\subsection{Datasets and Metrics}
Our method is trained and evaluated on two synthetic indoor datasets, ASE~\cite{avetisyan2024SceneScript} and Structured3D~\cite{zheng2020structured3d}, as well as the real-world SceneCAD dataset~\cite{avetisyan2020scenecad}. 

For layout estimation on ASE~\cite{avetisyan2024SceneScript}, we use $100$k scenes, randomly split into $95$k for training, $2.5$k for validation, and $2.5$k for testing. For object detection on ASE~\cite{avetisyan2024SceneScript}, bounding box annotations are not publicly available. We therefore adopt TR3D~\cite{DBLP:conf/icip/RukhovichVK23} from MMDetection3D~\cite{mmdet3d2020} to generate bounding box annotations by predicting bounding boxes on point clouds reconstructed from dense ground‑truth depth. We focus on the following object categories: cabinet, chair, table, sofa, and bed, which represent the most common objects in ASE. In total, $19$k scenes are used for training and $1$k scenes for evaluation. For Structured3D~\cite{zheng2020structured3d} and SceneCAD~\cite{avetisyan2020scenecad}, we use the official splits. 

To evaluate the accuracy of our method for layout estimation, we adopt the average per-class F1-Score ~\cite{DBLP:journals/corr/abs-2503-11806}. The F1-Score is computed at every $5$cm interval from $0$m to $1$m. Additionally, we also report the mean F1-Score across the categories. For object detection, we use F1-Score at $0.25$ and $0.5$ IoU thresholds following~\cite{avetisyan2024SceneScript}. To evaluate efficiency, we use three metrics: (1) \emph{Param} — the number of learnable parameters in the language decoder. (2) \emph{Latency} — the average decoding time for predicting layout for a scene, measured on NVIDIA RTX 2080 Ti. (3) $\alpha$ — the average number of accepted tokens per decoder inference.

\subsection{Implementation Details}
The Transformer decoder consists of $8$ layers, \ie, $4$ self-attention layers and $4$ cross-attention layers. Each layer operates on a feature dimension of $512$ with $8$ attention heads. The confidence head and the token head share a similar architecture, each consisting of three linear layers and two ReLU activations. The default value of $\tau$ is $2$. Our method is trained and evaluated under two settings, namely $n=8$ and $n=10$, as further increasing $n$ leads to performance degradation (see supplementary material). Additional details are provided in the supplementary material.

\begin{figure}[t]
  \centering
    \begin{subfigure}[t]{0.46\linewidth}
    \centering
    \includegraphics[width=\linewidth]{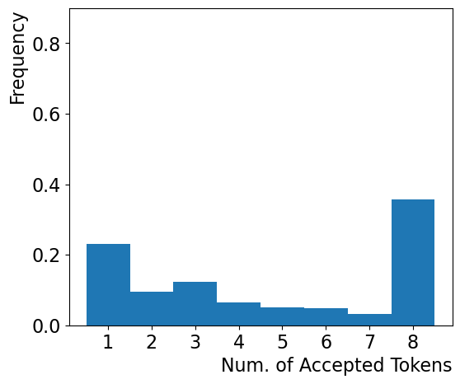}
    \caption{ProductThre}
    \label{fig:accepted_tokens_a}
  \end{subfigure}
  \hfill
  \begin{subfigure}[t]{0.46\linewidth}
    \centering
    \includegraphics[width=\linewidth]{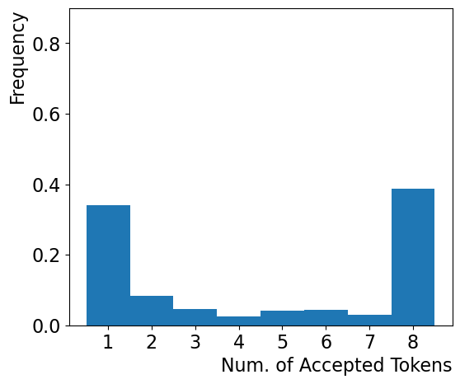}
    \caption{SoftmaxThre}
    \label{fig:accepted_tokens_b}
  \end{subfigure}

  \vspace{0.6em}

  \begin{subfigure}[t]{0.46\linewidth}
    \centering
    \includegraphics[width=\linewidth]{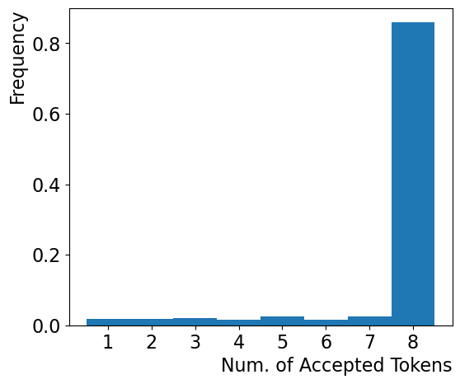}
    \caption{SSD}
    \label{fig:accepted_tokens_c}
  \end{subfigure}
  \hfill
  \begin{subfigure}[t]{0.46\linewidth}
    \centering
    \includegraphics[width=\linewidth]{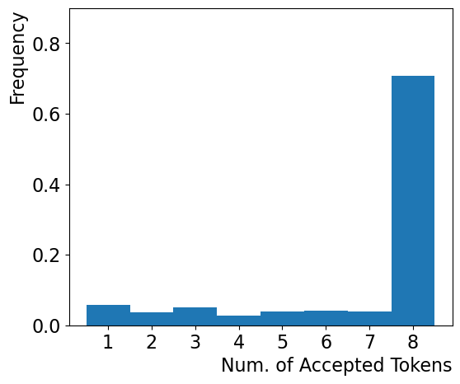}
    \caption{CGD}
    \label{fig:accepted_tokens_d}
  \end{subfigure}

   \caption{Distribution of the number of accepted tokens per decoder inference with different token filtering strategies. The distributions indicate that SSD and CGD in (c) and (d) accept a large number of tokens predicted by the network, whereas the ProductThre in (a) and SoftmaxThre in (b) reject many tokens. }
   \label{fig:accepted_tokens}
\end{figure}

\begin{figure}[t]
  \centering
    \begin{subfigure}[t]{0.94\linewidth}
    \centering
    \includegraphics[width=\linewidth]{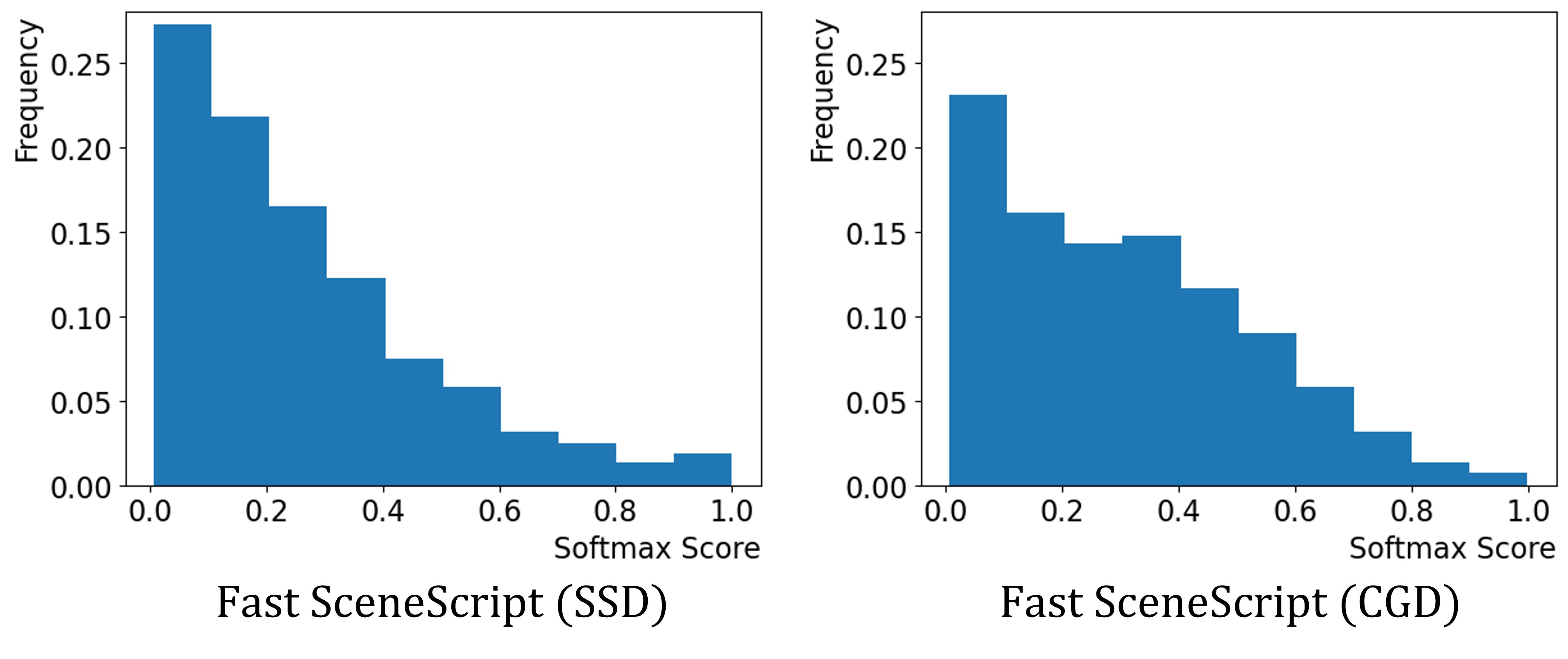}
    \caption{Softmax score distribution of the \textbf{first stopped} token}
    \label{fig:softmax_scores_a}
  \end{subfigure}

  \begin{subfigure}[t]{0.94\linewidth}
    \centering
    \includegraphics[width=\linewidth]{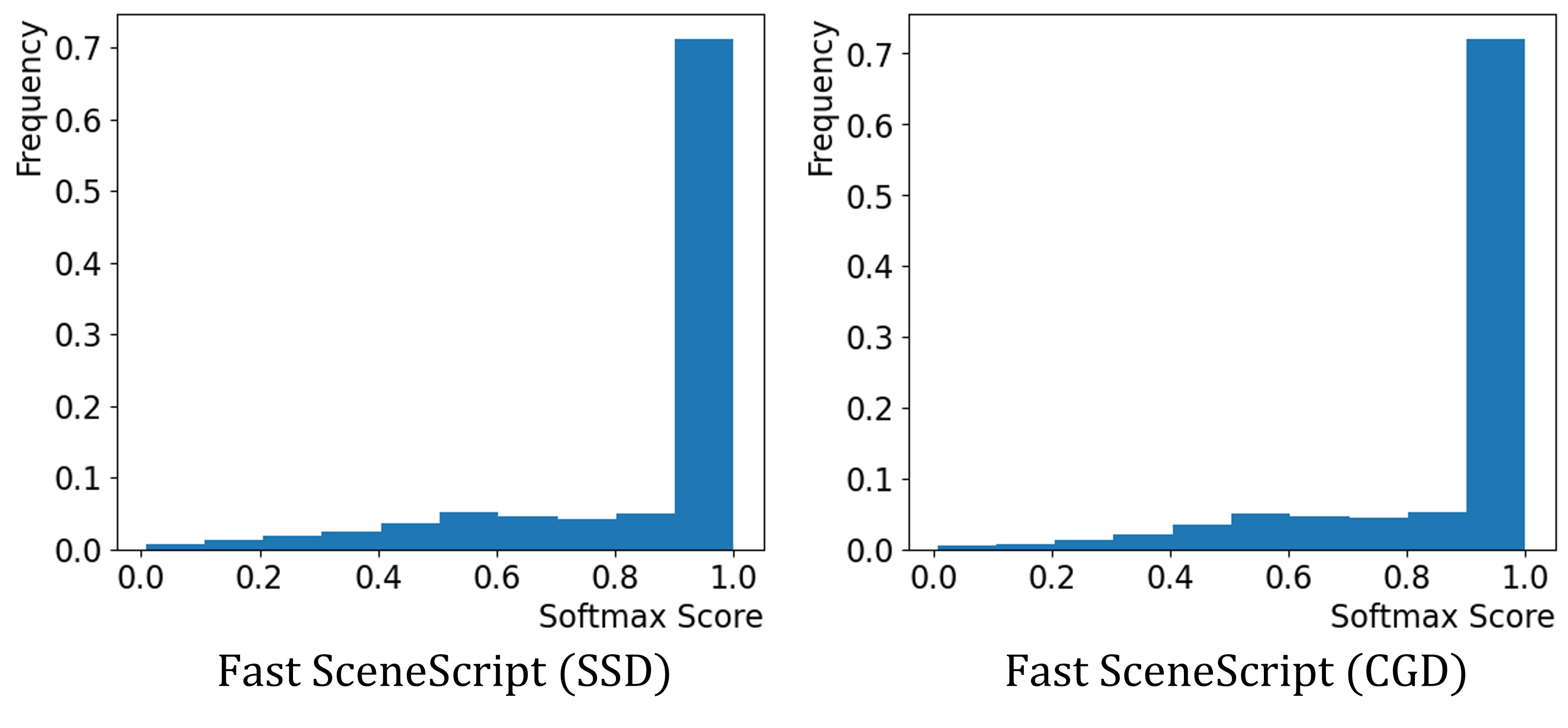}
    \caption{Softmax score distribution of \textbf{accepted} tokens}
    \label{fig:softmax_scores_b}
  \end{subfigure}

   \caption{Softmax score analysis for the first stopped and accepted tokens in Fast SceneScript. 
   The \emph{x}-axis shows the softmax scores and the \emph{y}-axis indicates their frequency on the ASE \emph{val} set~\cite{avetisyan2024SceneScript}.
   Our Fast SceneScript can stop when softmax confidence is high, yet can accept tokens even at low softmax confidence.
   }
   \label{fig:softmax_scores}
\end{figure}

\subsection{Layout Estimation}

This section evaluates the performance of our method for layout estimation. \cref{tab:results_aria}, ~\cref{tab:results_stru3d}, ~\cref{tab:sota_scenecad} report results on both \emph{val} and \emph{test} sets of ASE~\cite{avetisyan2024SceneScript},  Structured3D~\cite{zheng2020structured3d} \emph{test} set, and SceneCAD~\cite{avetisyan2020scenecad} \emph{val} set, respectively.

Compared to SceneScript~\cite{avetisyan2024SceneScript}, our Fast SceneScript achieves comparable or even better F1-Score with only a modest increase in parameters, while significantly speeding up inference. For instance, the comparison between Row \emph{i} and \emph{b} in~\cref{tab:results_aria} demonstrates that our Fast SceneScript is $5.09\times$ faster than SceneScript~\cite{avetisyan2024SceneScript} on ASE dataset. On Structured3D, the comparison between Row \emph{i} and \emph{b} in~\cref{tab:results_stru3d} shows that our Fast SceneScript is $5.57\times$ faster than SceneScript~\cite{avetisyan2024SceneScript} while even improving the mean F1-Score by $2.07\%$, consistent with previous findings in MTP~\cite{gloeckle2024better} and DeepSeek-V3~\cite{DBLP:journals/corr/abs-2412-19437} that the MTP auxiliary task acts as a beneficial regularizer. With CGD (Row \emph{j} in~\cref{tab:results_stru3d}), our method accelerates inference by $4.58\times$ over SceneScript while achieving a higher F1-Score, enabling efficient and effective on-the-fly decoding. Although SceneCAD \emph{train} set contains only around 1k scenes, which may limit the expressive power of a language model, our method in Row \emph{c} still achieves a $2.57\times$ efficiency gain over Row \emph{a} in ~\cref{tab:sota_scenecad}, showing clear and consistent improvements across synthetic and real-world scenarios.

SceneScript~\cite{avetisyan2024SceneScript} + MTP~\cite{gloeckle2024better} significantly degrades the accuracy compared to SceneScript~\cite{avetisyan2024SceneScript}. For instance, in~\cref{tab:results_aria}, the mean F1-Score of SceneScript + MTP with $10$ heads (Row \emph{h}) is 11.04\% lower than SceneScript (Row \emph{b}) on the \emph{test} set. Moreover, SceneScript + MTP increases the number of decoder parameters by $69.07\%$ and $88.79\%$ with $8$ and $10$ heads, respectively. In contrast, our Fast SceneScript achieves much better accuracy with fewer parameters. For example, in~\cref{tab:results_aria}, Fast SceneScript (Row \emph{i}) shows $12.04\%$ higher mean F1-Score than SceneScript + MTP (Row \emph{h}) on the \emph{test} set, while requiring $43.06\%$ fewer parameters. In ~\cref{tab:sota_scenecad}, Fast SceneScript in Row \emph{d} improves the F1-Score by $20$\%, compared to SceneScript + MTP in Row \emph{b}. These support the effectiveness of token filtering and parameter-efficient mechanisms.

We also compare our Fast SceneScript against the specialized layout estimation approach RoomFormer~\cite{yue2023connecting} in ~\cref{tab:results_stru3d}. We utilize the author-released model of RoomFormer for evaluation. Compared to RoomFormer~\cite{yue2023connecting}, our Fast SceneScript achieves a significantly higher F1-Score while using fewer decoder parameters. Although our Fast SceneScript is slower than RoomFormer, RoomFormer is limited to layout estimation only. In contrast, Fast SceneScript provides a general framework with potential applications to other 3D perception tasks, such as 3D object detection (see ~\cref{sec:det}) and 3D object part reconstruction.

Qualitative comparisons are presented in~\cref{fig:vis_aria}. Our Fast SceneScript produces results that closely align with the SceneScript~\cite{avetisyan2024SceneScript} and ground truth. Notably, our method generates more complete and accurate results compared to SceneScript~\cite{avetisyan2024SceneScript}+ MTP~\cite{gloeckle2024better}. Visualizations on Structured3D and SceneCAD are provided in the supplementary material.

\subsection{Object Detection}
\label{sec:det}

\begin{table}[h]
  \caption{Results of 3D object detection on ASE \emph{test} set (1k).}
  \label{tab:sota_aria_bbox}
  \footnotesize
  \centering
  \setlength{\tabcolsep}{2pt}
  \begin{tabular}{ll|cccc}
    \toprule

& Method    & Latency $\downarrow$  & $\alpha$ $\uparrow$ & F1 @.25 $\uparrow$ & F1 @.50 $\uparrow$ \\
     \midrule
a& SceneScript ~\cite{avetisyan2024SceneScript}  & 535 ms & 1 & 0.851 & 0.823 \\
    \midrule

 b&SceneScript ~\cite{avetisyan2024SceneScript} + MTP ~\cite{gloeckle2024better} & 83 ms & 8 & 0.815 & 0.772 \\
  c& Fast SceneScript (SSD)   &104 ms & 7.16 & 0.858  & 0.829\\
d& Fast SceneScript (CGD)  & 108 ms & 6.53 & 0.859  & 0.832\\
    \bottomrule
  \end{tabular}
\end{table}

We evaluate our method ($8$ heads) on 3D object detection using the ASE test set (1k scenes; \cref{tab:sota_aria_bbox}) and SceneCAD (\cref{tab:sota_scenecad}). Compared to SceneScript~\cite{avetisyan2024SceneScript}, our method (Row~\emph{c}) achieves a $5.14\times$ inference‑time speedup on ASE while maintaining accuracy, and a $3.35\times$ efficiency improvement with an $18\%$ accuracy gain on SceneCAD. Our method also consistently outperforms SceneScript + MTP, achieving, for example, a $7.38\%$ accuracy gain with CGD on ASE.

\subsection{Ablation Experiments}

\cref{tab:ab_aria} presents the ablation study of our method on ASE \emph{val} set~\cite{avetisyan2024SceneScript} for layout estimation with $n=8$. Rows \emph{a, b} demonstrate that MTP can accelerate the inference but sacrifices accuracy. In contrast, Fast SceneScript achieves a better trade-off among inference speed, accuracy, and parameter efficiency (see Rows \emph{d, i}).

For SSD-based methods, Rows \emph{c} - \emph{e} illustrate that incorporating loosened criterion ~\cref{eq:distance_metric} for numerical tokens significantly improves efficiency, as evaluations with $\tau=2$ and $\tau=5$ are faster than with $\tau=0$. For both efficiency and effectiveness, we finalize the setting at $\tau=2$.

We compare proposed scoring strategy, CGD, with two common baselines: (1) \emph{ProductThre}~\cite{DBLP:conf/naacl/TuliLHJSJ24, DBLP:journals/corr/abs-2509-24007} - assesses token reliability using the cumulative product of softmax scores, (2) \emph{SoftmaxThre}~\cite{DBLP:journals/corr/abs-2502-09992} - based on the softmax score of token logits. The thresholds for \emph{ProductThre} and \emph{SoftmaxThre} are set to 0.3 and 0.5, respectively, for efficiency–accuracy trade‑offs. Our CGD (Row \emph{i}) is significantly more efficient than \emph{ProductThre} (Row \emph{g}) and \emph{SoftmaxThre} (Row \emph{h}). As shown in \cref{fig:accepted_tokens}, CGD accepts more predicted tokens, while \emph{ProductThre} and \emph{SoftmaxThre} reject many due to low confidence in challenging tokens (e.g., windows). See supplementary material for details. Moreover, \cref{fig:softmax_scores} shows that our Fast SceneScript can reject tokens with high softmax scores and also accept tokens with low softmax scores. This demonstrates that the softmax score can not accurately reflect the actual token reliability.

The comparisons between Row \emph{d} and Row \emph{f}, as well as Row \emph{i} and Row \emph{j}, indicate that our parameter-efficient mechanism reduces the number of parameters without compromising accuracy. Notably, the decoder in Row \emph{i} contains $48.11\%$ fewer parameters than that in Row \emph{j}.

\section{Conclusion}
\label{sec:conclusion}

This work proposes Fast SceneScript, a novel, accurate, and efficient architecture for 3D scene understanding, including layout estimation and object detection. We apply multi-token prediction (MTP) to accelerate the inference. To avoid the accuracy drop of MTP, two types of decoding with token filtering strategies are investigated, \ie, SSD and CGD. Each decoding strategy offers a distinct balance between flexibility and performance: SSD enables higher token throughput per inference, while CGD provides on-the-fly decoding. Furthermore, parameter-efficient mechanism is introduced, which can reduce the parameters by $43\%$ compared to MTP. Experimental results on both synthetic and real-world datasets have demonstrated the efficiency and effectiveness of our method.

\section*{Acknowledgements}
\label{sec:acknowledgements}

The authors thank Alin Dondera for data preprocessing and for providing the evaluation metrics for the object detection task. The authors also thank Jihong Ju for proofreading.
{
    \small
    \bibliographystyle{ieeenat_fullname}
    \bibliography{main}
}

\clearpage
\setcounter{page}{1}
\maketitlesupplementary

\setcounter{section}{0}
\setcounter{figure}{0}
\setcounter{table}{0}
\renewcommand{\thefigure}{S\arabic{figure}}
\renewcommand{\thetable}{S\arabic{table}}
\renewcommand{\thesection}{\Alph{section}}

\cref{sec:extra_result} provides additional comparisons, along with extended ablation studies, distribution analyses, and failure cases. Additional implementation details are provided in~\cref{sec:imple_details}. Additional related work is discussed in~\cref{sec:supp_related_works}. Details of the ASE dataset splits are given in~\cref{sec:data_split}.

\begin{table*}
  \caption{Quantitative comparisons for layout estimation on ASE dataset~\cite{avetisyan2024SceneScript}. $n$ denotes the number of MTP heads. $\alpha_{val}$ and $\alpha_{test}$ refer to the average number of tokens accepted per decoder inference on \emph{val} and \emph{test} sets.
  }
  \label{tab:sota_aria_supp}
  \small
  \centering
  \setlength{\tabcolsep}{3pt}
  \begin{tabular}{lccc|c|cccc|c|cccc}
    \toprule

 \multirow{2}{*}{Method} & \multirow{2}{*}{$n$} & \multirow{2}{*}{Param $\downarrow$} & \multirow{2}{*}{Latency $\downarrow$} & \multirow{2}{*}{$\alpha_{val}$ $\uparrow$} & \multicolumn{4}{c|}{F1-Score of \emph{val} set $\uparrow$} & \multirow{2}{*}{$\alpha_{test}$ $\uparrow$} & \multicolumn{4}{c}{F1-Score of \emph{test} set $\uparrow$} \\
 & & & & & wall & window & door & mean & & wall  & window  & door & mean\\
     \midrule
    SceneScript~\cite{avetisyan2024SceneScript} & 1 & 14.00 M & 382 ms &1 & 0.918 & 0.880 & 0.940 & 0.913 & 1 & 0.921 & 0.881  & 0.942 & 0.915 \\
    \midrule
    SceneScript~\cite{avetisyan2024SceneScript} + MTP~\cite{gloeckle2024better} & 8 & 23.67 M  & 62 ms & 8& 0.831 & 0.804 & 0.885 & 0.840 &  8  & 0.836  & 0.804 & 0.886 & 0.842\\
    Fast SceneScript (SSD) & 8  & 15.05 M & 81 ms & 7.46 & 0.914 &0.882 & 0.939 & 0.912 & 7.45  & 0.919  & 0.882 & 0.939 & 0.913\\
    Fast SceneScript (CGD) & 8 & 16.10 M & 92 ms &6.29 & 0.912 & 0.883 & 0.938 &0.911 & 6.30 & 0.918 & 0.883  & 0.938 & 0.913 \\
    \midrule
  
   SceneScript~\cite{avetisyan2024SceneScript} + MTP~\cite{gloeckle2024better} & 10 & 26.43 M  & 54 ms &10 & 0.805 & 0.776 & 0.863 & 0.815 & 10  & 0.808 & 0.774 & 0.861 & 0.814\\
    Fast SceneScript (SSD) & 10 & 15.05 M &75 ms& 8.97 & 0.910 & 0.879 & 0.937 & 0.909&  8.99 & 0.915 & 0.880 &  0.940 & 0.912\\
    Fast SceneScript (CGD) & 10 & 16.10 M & 89 ms & 7.27& 0.909 & 0.880 & 0.936 & 0.908& 7.27 & 0.912  & 0.879  & 0.938 & 0.910 \\
    \midrule

    SceneScript~\cite{avetisyan2024SceneScript} + MTP~\cite{gloeckle2024better} & 12 &  29.20 M  & 49 ms & 12 & 0.792 & 0.754 & 0.840 & 0.795 &   12 & 0.800  & 0.752 & 0.842 & 0.798\\
    Fast SceneScript (SSD) & 12  &  15.05 M &  75 ms & 10.21 & 0.902 &0.877 & 0.934 & 0.904 & 10.11  & 0.907  & 0.876 & 0.937 & 0.907 \\
    Fast SceneScript (CGD) & 12 & 16.10 M & 93 ms & 7.71 & 0.902 & 0.877 & 0.933 &0.904 & 7.86 & 0.905 & 0.877  & 0.937 & 0.906 \\
    \midrule
     SceneScript~\cite{avetisyan2024SceneScript} + MTP~\cite{gloeckle2024better} & 16 &  34.72 M  & 42 ms & 16 & 0.717 & 0.681 & 0.794 & 0.731 &  16  & 0.721  & 0.688 & 0.794 & 0.734 \\
    Fast SceneScript (SSD) & 16  & 15.05 M & 73 ms & 12.48 & 0.898 &0.870 & 0.932 & 0.900 & 12.38  & 0.902  & 0.870 & 0.933 & 0.902\\
    Fast SceneScript (CGD) & 16 & 16.10 M &  99 ms & 8.60 & 0.896 & 0.874 & 0.934 &0.901 & 8.65 & 0.898 & 0.871  & 0.934 & 0.901 \\

    \bottomrule
  \end{tabular}
\end{table*}

\begin{table*}
  \caption{Ablation experiments for layout estimation on the ASE \emph{val} set~\cite{avetisyan2024SceneScript}. $\epsilon$ is the threshold used to determine where to stop in the scoring-based methods. The default setting in our paper is \underline{underlined}. 
  }
  \label{tab:ab_aria_thre_supp}
  \small
  \centering
  \setlength{\tabcolsep}{5pt}
  \begin{tabular}{lccccc|cccc}
    \toprule

\multirow{2}{*}{Method} & \multirow{2}{*}{$\epsilon$}  & \multirow{2}{*}{$n$} & \multirow{2}{*}{Param $\downarrow$} & \multirow{2}{*}{Latency $\downarrow$} & \multirow{2}{*}{$\alpha_{val}$ $\uparrow$} & \multicolumn{4}{c}{F1-Score $\uparrow$}\\
  &  &  & & & & wall & window & door & mean \\

     \midrule

 Fast SceneScript (CGD) & 0.80& 8 & 16.10 M&  85 ms& 6.77 & 0.905 & 0.878 & 0.935 & 0.906 \\
 Fast SceneScript (CGD) & 0.85& 8 & 16.10 M&  88 ms& 6.55 & 0.910 & 0.879 & 0.937 & 0.909 \\
 Fast SceneScript (CGD) & \underline{0.90} & 8 & 16.10 M &  91 ms & 6.29 & 0.912 & 0.883  & 0.938 & 0.911 \\
 Fast SceneScript (CGD) & 0.95& 8 & 16.10 M&  97 ms& 5.98 & 0.914 & 0.883 & 0.939 & 0.912 \\ \midrule

 Fast SceneScript (ProductThre~\cite{DBLP:conf/naacl/TuliLHJSJ24, DBLP:journals/corr/abs-2509-24007}) & 0.20& 8 & 16.10 M& 97 ms& 5.42 & 0.905 & 0.878 & 0.934 & 0.907 \\
 Fast SceneScript (ProductThre~\cite{DBLP:conf/naacl/TuliLHJSJ24, DBLP:journals/corr/abs-2509-24007}) & 0.25 & 8 & 16.10 M&  105 ms&5.04  & 0.907 & 0.880 & 0.938 & 0.908 \\
 Fast SceneScript (ProductThre~\cite{DBLP:conf/naacl/TuliLHJSJ24, DBLP:journals/corr/abs-2509-24007}) & \underline{0.30} & 8 & 16.10 M& 112 ms& 4.67 & 0.908 & 0.881 & 0.938 & 0.909 \\
 Fast SceneScript (ProductThre~\cite{DBLP:conf/naacl/TuliLHJSJ24, DBLP:journals/corr/abs-2509-24007}) & 0.35 & 8 & 16.10 M& 120 ms& 4.34 & 0.909 & 0.882 & 0.938 & 0.910 \\ \midrule
 Fast SceneScript (SoftmaxThre~\cite{DBLP:journals/corr/abs-2502-09992}) & 0.40 & 8 & 16.10 M& 96 ms& 5.49 & 0.903 & 0.875 & 0.931 & 0.903 \\
 Fast SceneScript (SoftmaxThre~\cite{DBLP:journals/corr/abs-2502-09992}) & 0.45 & 8 & 16.10 M& 106 ms& 5.03 & 0.906 & 0.878 & 0.935 & 0.906 \\
 Fast SceneScript (SoftmaxThre~\cite{DBLP:journals/corr/abs-2502-09992}) & \underline{0.50} & 8 & 16.10 M&  116 ms&4.53  & 0.908 & 0.880 & 0.937 & 0.908 \\
 Fast SceneScript (SoftmaxThre~\cite{DBLP:journals/corr/abs-2502-09992}) & 0.55 & 8 & 16.10 M& 131 ms& 3.57 & 0.912 & 0.882 & 0.939 & 0.911 \\
    \bottomrule
  \end{tabular}
\end{table*}

\begin{figure*}[htbp]
  \centering
  \includegraphics[width=0.9\linewidth]{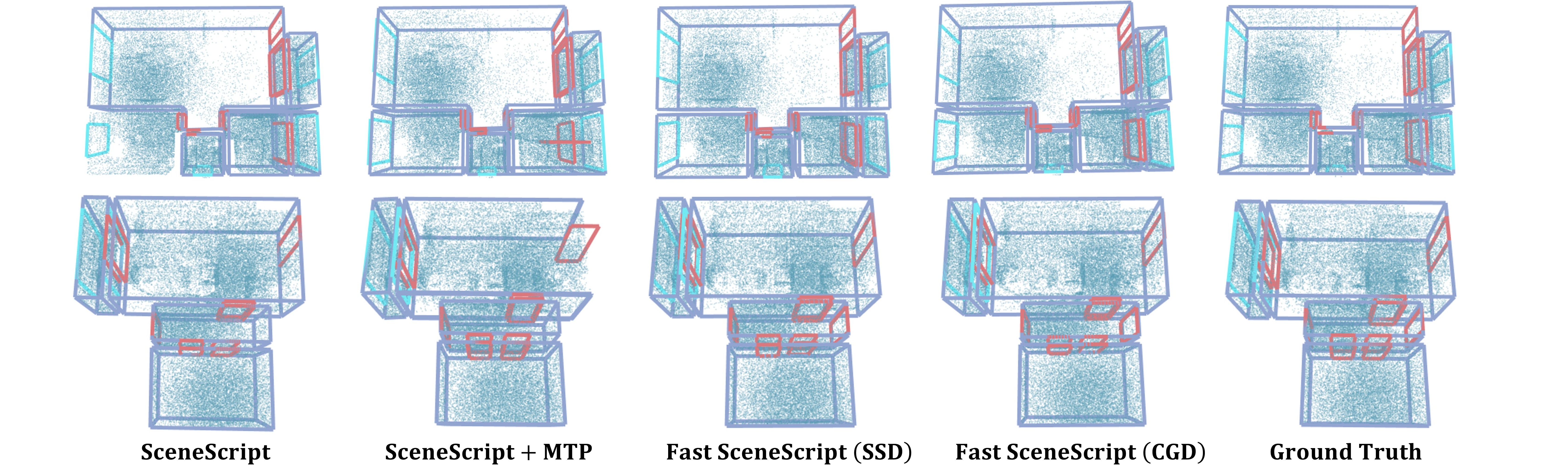}

   \caption{Qualitative results on Structured3D \emph{test} set~\cite{zheng2020structured3d}. The number of MTP heads $n$ is set to 8. The scene layout generated by Fast SceneScript demonstrates superior accuracy compared to SceneScript~\cite{avetisyan2024SceneScript} + MTP~\cite{gloeckle2024better}.}
   \label{fig:vis_stru3d}
\end{figure*}

\begin{figure*}
  \centering
\begin{subfigure}[t]{0.78\linewidth}
    \centering
    \includegraphics[width=\linewidth]{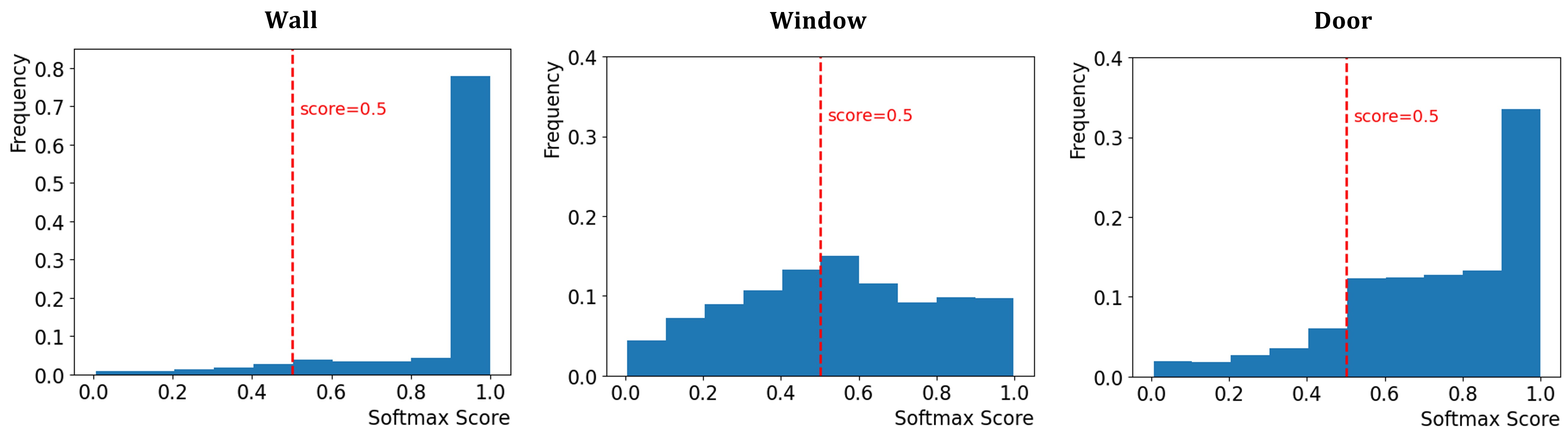}
    \caption{Fast SceneScript (without Token Filtering)}
  \end{subfigure}

  \begin{subfigure}[t]{0.78\linewidth}
    \centering
    \includegraphics[width=\linewidth]{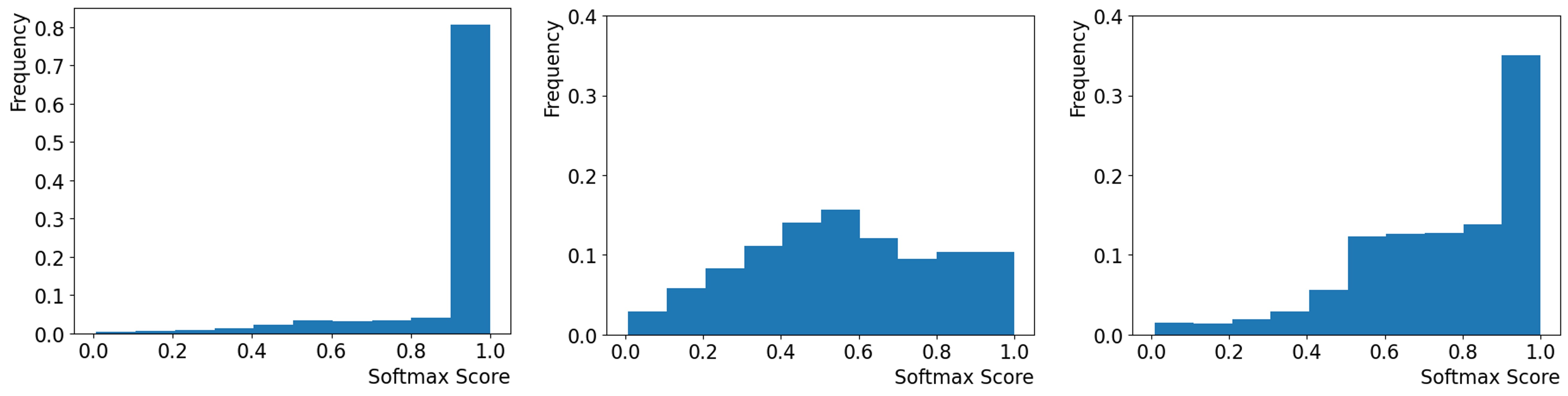}
    \caption{Fast SceneScript (SSD)}
  \end{subfigure}

  \begin{subfigure}[t]{0.78\linewidth}
    \centering
    \includegraphics[width=\linewidth]{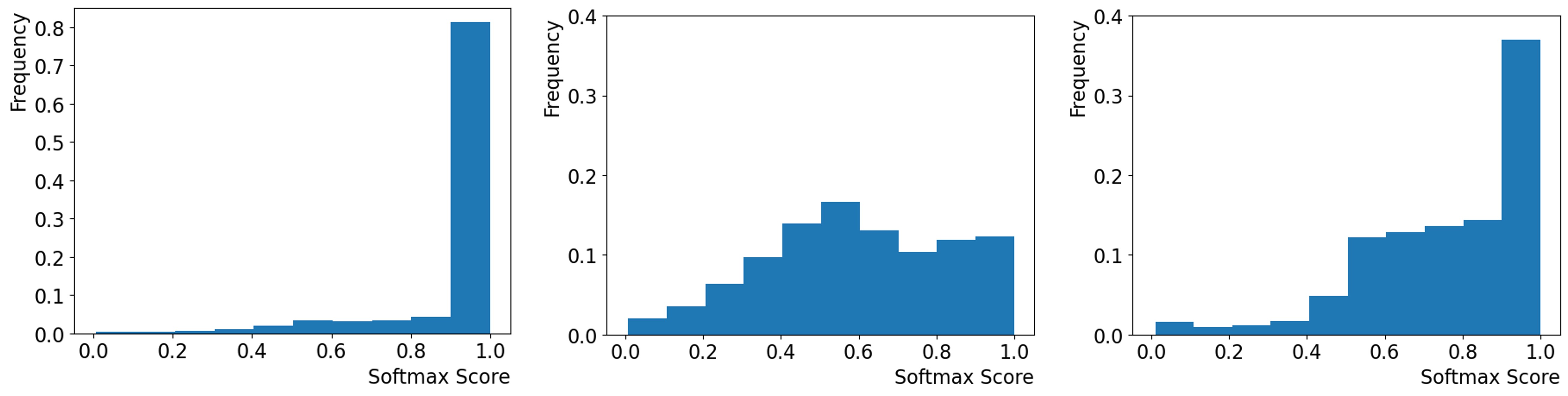}
    \caption{Fast SceneScript (CGD)}
  \end{subfigure}

   \caption{Softmax score distribution of $3$ classes. (a): softmax score distributions of all tokens without filtering. (b) - (c): softmax score distributions for accepted tokens. In (a), the red dashed line marks the position where the softmax score equals $0.5$. Tokens to the left of this line are rejected by \emph{SoftmaxThre}. In contrast, The SSD in (b) and CGD in (c) can accept tokens with low softmax confidence.  }
   \label{fig:supp_dis}
\end{figure*}

\section{Additional Results}
\label{sec:extra_result}

\subsection{Results with More Heads}
\label{sec:result_ase_supp}
\cref{tab:sota_aria_supp} illustrates the impact of using more heads in our Fast SceneScript for layout estimation. We observe a drop in accuracy and no obvious latency advantage when using $12$ or $16$ heads in Fast SceneScript.

\subsection{Qualitative Results on Structured3D}
\cref{fig:vis_stru3d} shows qualitative results on Structured3D~\cite{zheng2020structured3d}. Compared to SceneScript and SceneScript + MTP, our Fast SceneScript yields more complete and accurate layouts.

\subsection{Ablation Study for Layout Estimation}
\label{sec:ab_ase_supp}
This section presents additional ablation study on ASE~\cite{avetisyan2024SceneScript}.

{\bf $\epsilon$ for the scoring-based methods:} \cref{tab:ab_aria_thre_supp} reports results of scoring-based token filtering methods under different thresholds $\epsilon$ during inference. The results indicate minor variations across different thresholds. To balance accuracy and speed, we set the default thresholds to $0.90$, $0.30$, and $0.50$ for CGD, ProductThre, and SoftmaxThre, respectively.

{\bf $\tau$ for the supervision of confidence branch:}  \cref{tab:ab_aria_supp} illustrates the results of CGD in our Fast SceneScript trained with different values of $\tau$, \ie, $0$, $2$, $5$. Among these, Fast SceneScript (CGD) with $\tau=2$ achieves the highest F1-Score and is therefore selected as the final configuration.

{\bf Softmax score distribution:} 
\cref{fig:supp_dis} provides detailed softmax score distribution for $3$ classes: wall, window, and door. In (a), for \emph{window}, many tokens generated by the MTP model have softmax scores below $0.5$. As a result, the method with \emph{SoftmaxThre} rejects these tokens. In contrast, the SSD in (b) and CGD in (c) also accept some of these tokens. This shows the effectiveness of the proposed CGD strategy compared to naive decoding using softmax probabilities as the reliability of a token.

{\bf Supervision strategies for confidence branch:} \cref{tab:ab_aria_gt_supp} investigates the impact of different training losses for confidence branch in CGD. Two settings are considered: (1) Generating the confidence label based on the consistency with the token ${t}_{k+i}^1$ from the first head (our default setting, \emph{Fast SceneScript (CGD)}), (2) Generating the confidence label based on the consistency with its ground truth, \ie, \emph{Fast SceneScript (CGD, GT)}. While both strategies achieve comparable accuracy, the default setting is faster than \emph{Fast SceneScript (CGD, GT)}. We argue that our default setting not only learns token confidence but also captures the inherent uncertainty among different heads. By comparing with predictions rather than with ground truth, the confidence branch is encouraged to model internal agreement and disagreement, which also represents the model uncertainty. This makes confidence estimation more robust and informative. As a result, the network accept more tokens during inference, reaching faster speed.

\subsection{Analyses for Object Detection}
This section provides additional analysis for object detection.

\begin{figure}
  \centering
  \begin{subfigure}[t]{0.46\linewidth}
    \centering
    \includegraphics[width=\linewidth]{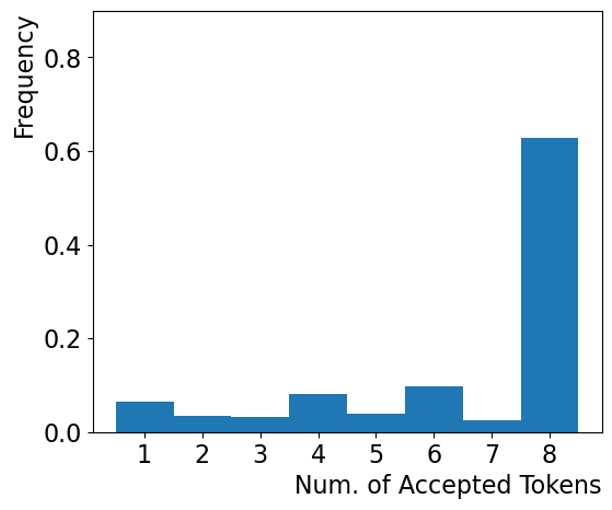}
    \caption{CGD}
    \label{fig:accepted_token_det_a}
  \end{subfigure}
  \hfill
  \begin{subfigure}[t]{0.46\linewidth}
    \centering
    \includegraphics[width=\linewidth]{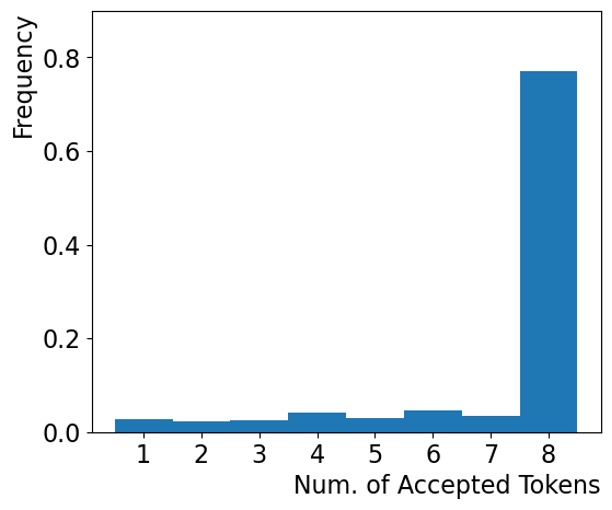}
    \caption{SSD}
    \label{fig:accepted_token_det_b}
  \end{subfigure}

   \caption{Distribution of the number of accepted tokens per decoder inference with our Fast SceneScript for object detection. It can be seen that both SSD and CGD accept a large number of tokens predicted by the network.}
   \label{fig:accepted_token_det}
\end{figure}

{\bf Distribution of accepted tokens per inference:}
\cref{fig:accepted_token_det} reports the distribution of the number of accepted tokens per decoder inference with our Fast SceneScript for object detection. It shows that our Fast SceneScript with CGD or SSD accepts all $8$ tokens in $\sim 70\%$ of decoder runs. This aligns with our observation on the layout estimation task.

\begin{figure}[htbp]
  \centering
  \begin{subfigure}[t]{0.46\linewidth}
    \centering
    \includegraphics[width=\linewidth]{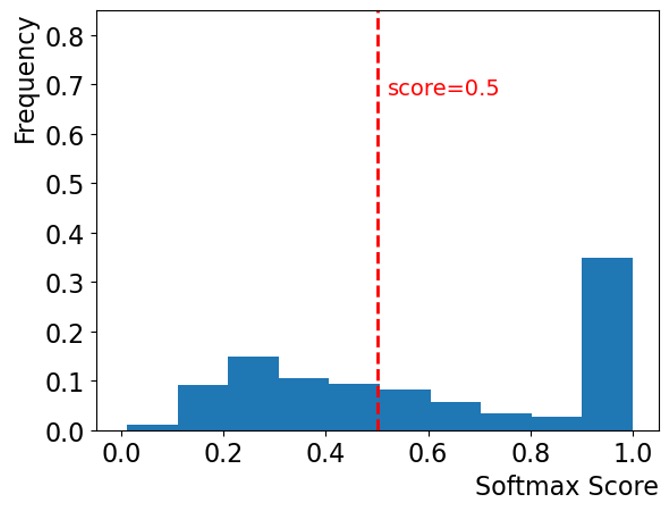}
    \caption{CGD}
    \label{fig:softmax_scores_det_a}
  \end{subfigure}
  \hfill
  \begin{subfigure}[t]{0.46\linewidth}
    \centering
    \includegraphics[width=\linewidth]{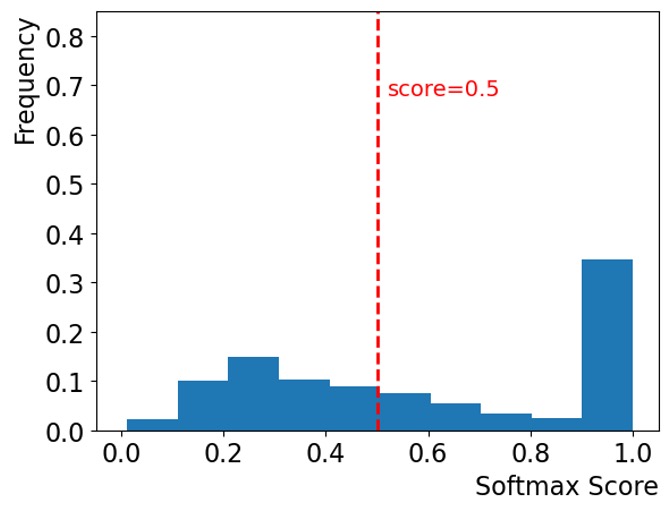}
    \caption{SSD}
    \label{fig:softmax_scores_det_b}
  \end{subfigure}

   \caption{Analysis of softmax scores for the accepted tokens in Fast SceneScript for object detection. The \emph{x}-axis shows the softmax scores and the \emph{y}-axis indicates their frequency. Our Fast SceneScript also accepts tokens with low softmax scores. }
   \label{fig:softmax_scores_det}
\end{figure}

{\bf Softmax score distribution:}
\cref{fig:softmax_scores_det} shows the softmax score distribution of the accepted tokens in Fast SceneScript. Only $\sim 40\%$ of accepted tokens have scores above $0.9$, compared to $\sim70\%$ in layout estimation (see main paper). We attribute the high softmax confidence in layout estimation to inherent token redundancy and predictable layout structure, \eg, neighboring walls often share a corner. In contrast, there is no redundancy in 3D object detection tokens, yielding lower softmax scores.

\subsection{Failure Cases}

\begin{figure*}
  \centering
  \begin{subfigure}{0.4\linewidth}
    \centering
    \includegraphics[width=\linewidth]{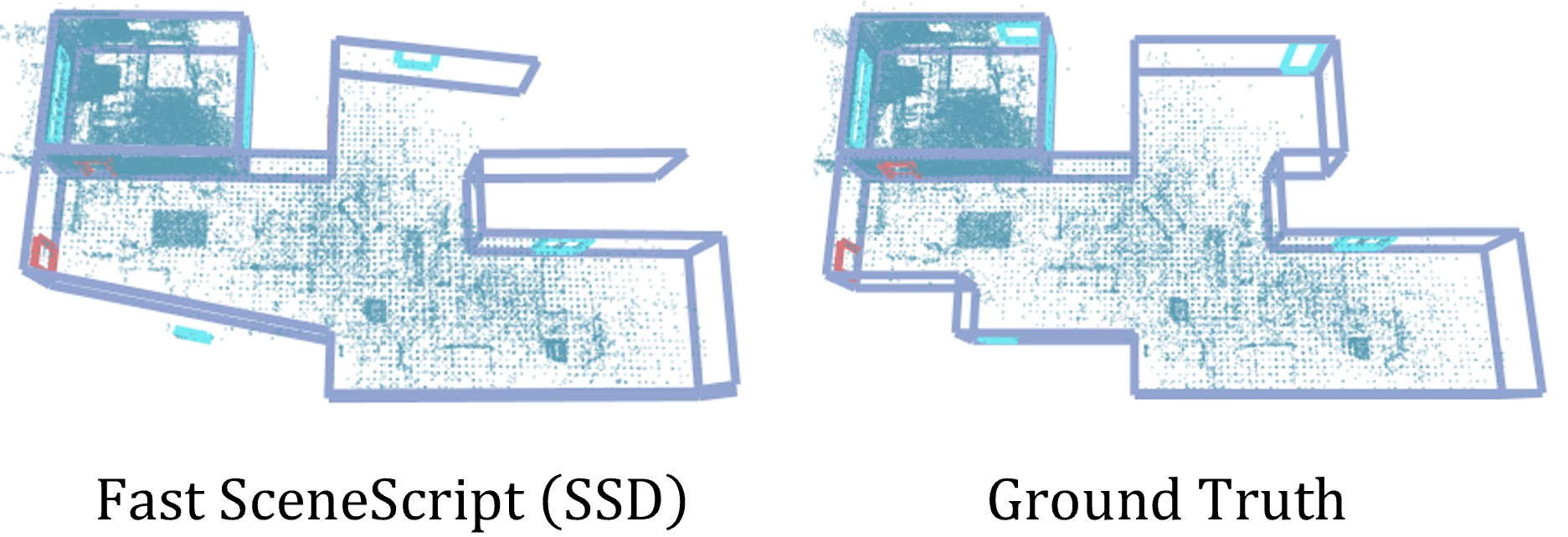}
    \caption{ASE~\cite{avetisyan2024SceneScript} (synthetic)}
  \end{subfigure} \hspace{0.05\linewidth}
  \begin{subfigure}{0.38\linewidth}
    \centering
    \includegraphics[width=\linewidth]{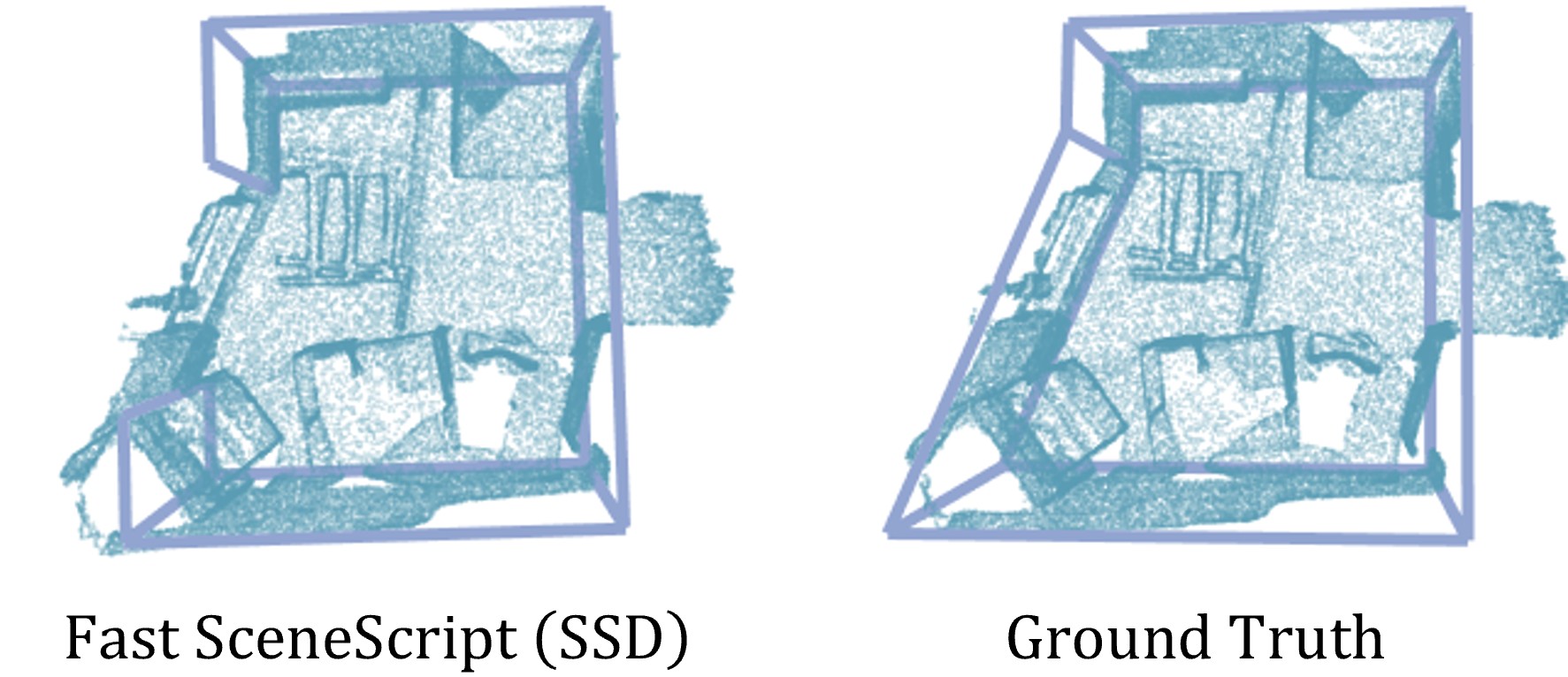}
    \caption{SceneCAD~\cite{avetisyan2020scenecad} (real-world)}
  \end{subfigure}
   \caption{Failure cases on ASE~\cite{avetisyan2024SceneScript} and SceneCAD~\cite{avetisyan2020scenecad} datasets.}
   \label{fig:failure}
\end{figure*}

\cref{fig:failure} presents failure cases of our method on the synthetic ASE~\cite{avetisyan2024SceneScript} and real-world SceneCAD~\cite{avetisyan2020scenecad} datasets. It can be seen that Fast SceneScript (SSD) may struggle in regions where the point cloud is incomplete and in producing accurate layouts for certain non-Manhattan scenes.

\subsection{SceneScript with Longer Training}

\cref{tab:sota_aria_baseline} illustrates the results of the baseline SceneScript~\cite{avetisyan2024SceneScript} trained for $60$ and $90$ epoches on layout estimation. Longer training yields marginal improvement in accuracy.

\begin{table*}
  \caption{Ablation study for layout estimation on ASE \emph{val} set~\cite{avetisyan2024SceneScript}. The default setting in our paper is \underline{underlined}.}
  \label{tab:ab_aria_supp}
  \small
  \centering
  \setlength{\tabcolsep}{5pt}
  \begin{tabular}{lccccc|cccc}
    \toprule

 \multirow{2}{*}{Method}  & \multirow{2}{*}{$\tau$} & \multirow{2}{*}{$n$} & \multirow{2}{*}{Param $\downarrow$} & \multirow{2}{*}{Latency $\downarrow$} & \multirow{2}{*}{$\alpha_{val}$ $\uparrow$} & \multicolumn{4}{c}{F1-Score of \emph{val} set $\uparrow$}  \\
 & & & & & & wall & window & door & mean \\
     \midrule

 Fast SceneScript (CGD) &0 & 8 & 16.10 M  & 97 ms & 5.94 & 0.908 & 0.881 & 0.935 & 0.908  \\ 
     
  Fast SceneScript (CGD) & \underline{2} & 8 & 16.10 M & 91 ms & 6.29 & 0.912 & 0.883 & 0.938 &0.911 \\

 Fast SceneScript (CGD) & 5& 8 & 16.10 M  & 90 ms & 6.44 & 0.912 & 0.882 & 0.936 & 0.910\\ 

    \bottomrule
  \end{tabular}
\end{table*}

\begin{table*}
  \caption{Ablation experiments for layout estimation on the ASE \emph{val} set~\cite{avetisyan2024SceneScript}. 
  }
  \label{tab:ab_aria_gt_supp}
  \small
  \centering
  \setlength{\tabcolsep}{5pt}
  \begin{tabular}{lcccc|cccc}
    \toprule

   \multirow{2}{*}{Method} & \multirow{2}{*}{$n$} & \multirow{2}{*}{Param $\downarrow$} & \multirow{2}{*}{Latency $\downarrow$} & \multirow{2}{*}{$\alpha_{val}$ $\uparrow$} & \multicolumn{4}{c}{F1-Score $\uparrow$}\\
  &  &  &  & & wall & window & door & mean \\ \midrule

   Fast SceneScript (CGD, GT) & 8 & 16.10 M& 100 ms& 5.61 & 0.913 & 0.884 & 0.935 & 0.911 \\
   Fast SceneScript (CGD) & 8 & 16.10 M &  91 ms & 6.29 & 0.912 & 0.883  & 0.938 & 0.911 \\
 
    \bottomrule
  \end{tabular}
\end{table*}

\begin{table*}
  \caption{Results of the baseline SceneScript~\cite{avetisyan2024SceneScript} with longer training on ASE dataset~\cite{avetisyan2024SceneScript}. 
  }
  \label{tab:sota_aria_baseline}
  \small
  \centering
  \setlength{\tabcolsep}{7pt}
  \begin{tabular}{lc|cccc|cccc}
    \toprule

 \multirow{2}{*}{Method} & \multirow{2}{*}{Epoch}  & \multicolumn{4}{c|}{F1-Score of \emph{val} set $\uparrow$} & \multicolumn{4}{c}{F1-Score of \emph{test} set $\uparrow$} \\
 & & wall & window & door & mean & wall  & window  & door & mean\\
     \midrule

    SceneScript~\cite{avetisyan2024SceneScript} & 60 & 0.918 & 0.880 & 0.940 & 0.913 & 0.921 & 0.881  & 0.942 & 0.915 \\
    SceneScript~\cite{avetisyan2024SceneScript} & 90 & 0.921 & 0.887 & 0.946 & 0.918 & 0.924 & 0.888  & 0.948 & 0.920 \\
    
    \bottomrule
  \end{tabular}
\end{table*}

\begin{figure*}[!htbp]
  \centering
  \includegraphics[width=1.0\linewidth]{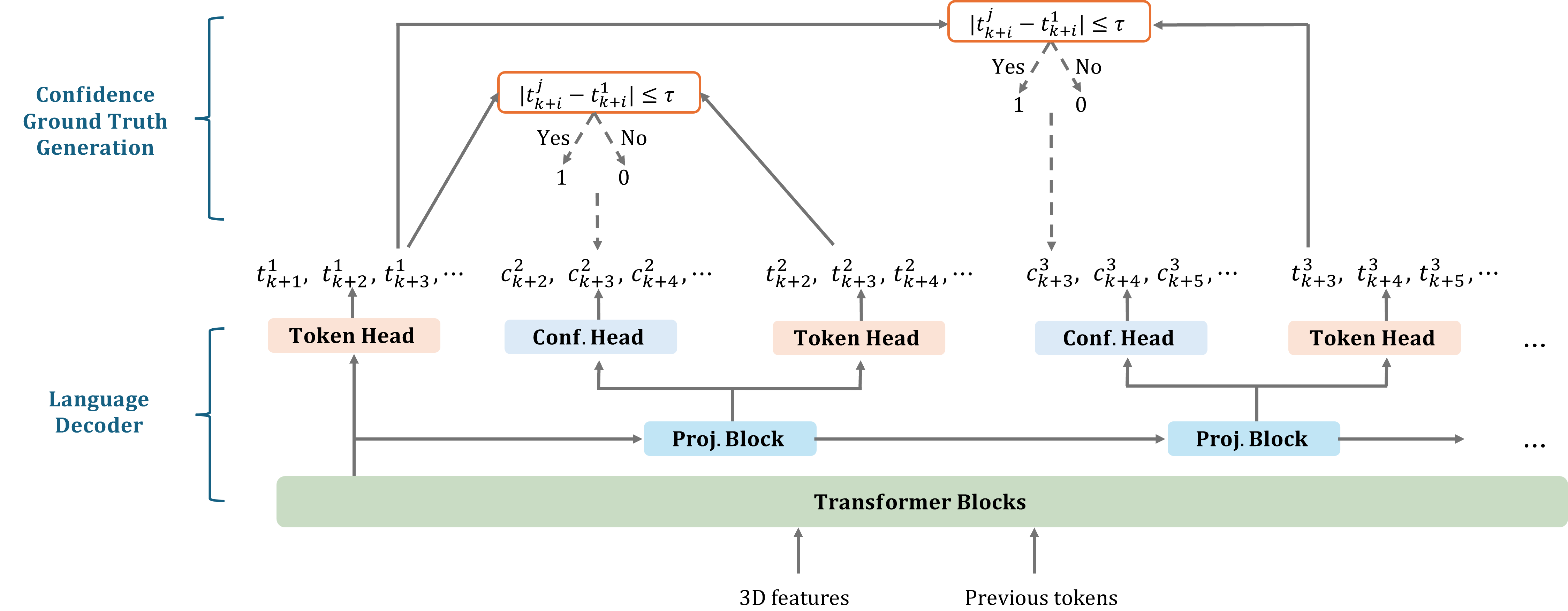}

   \caption{Confidence ground truth generation for CGD. During training, the confidence label $\hat{c}_{k+i}^j$ for the $(k+i)$-th token ${t}_{k+i}^j$, generated by the $j$-th head ($j \in [2, n]$), is computed as follows: (1) Calculate the absolute difference between the token from head $j$ and the token from the first head, \ie, $|{t}_{k+i}^j - {t}_{k+i}^1|$, (2) Apply thresholding: if $|{t}_{k+i}^j - {t}_{k+i}^1| \leq \tau$, $\hat{c}_{k+i}^j=1$, otherwise, $\hat{c}_{k+i}^j=0$. In particular, $\tau$ is a positive hyperparameter for numerical tokens and is 0 for non-numerical tokens. This figure illustrates examples of generating the confidence labels $\hat{c}_{k+3}^2$ and $\hat{c}_{k+3}^3$.}
   \label{fig:supp_cgd_training}
\end{figure*}

\begin{figure*}[!htbp]
  \centering
  \begin{subfigure}[t]{0.3\linewidth}
    \centering
    \includegraphics[width=\linewidth]{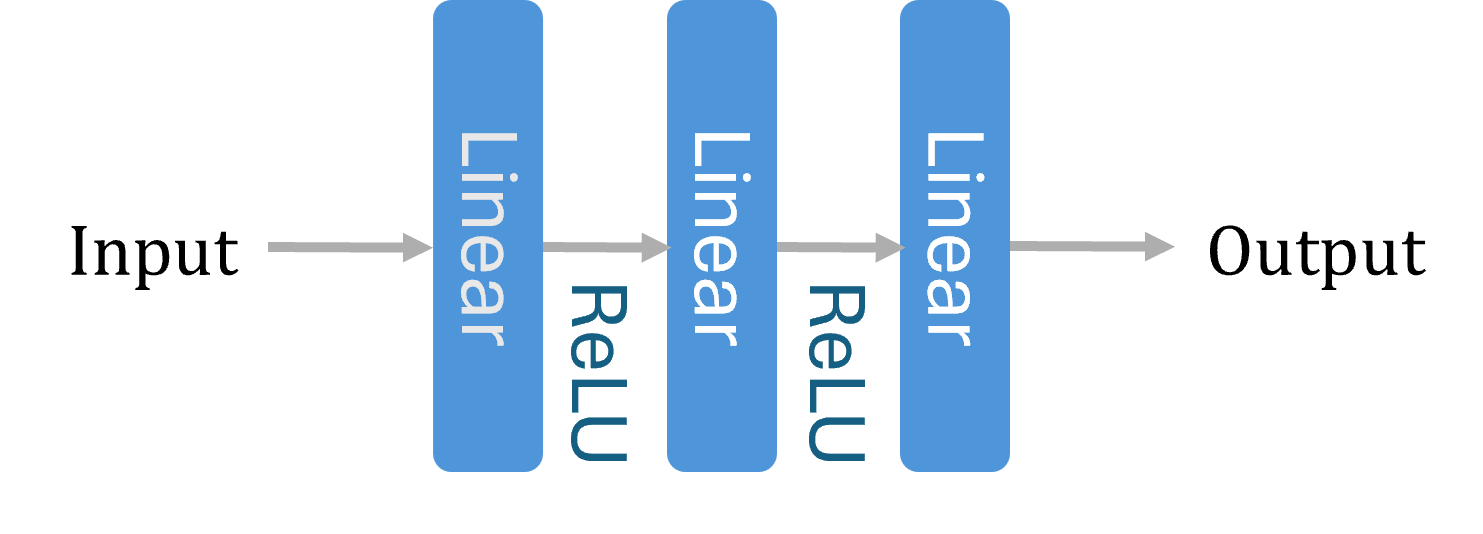}
    \caption{Token / Confidence Head}
    \label{fig:supp_details_head}
  \end{subfigure}
  \hfill
  \begin{subfigure}[t]{0.55\linewidth}
    \centering
    \includegraphics[width=\linewidth]{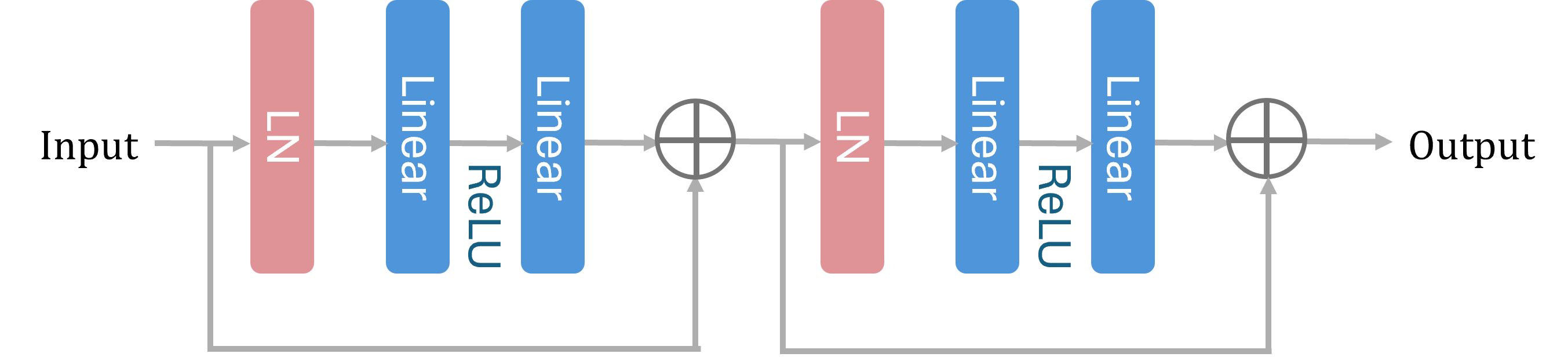}
    \caption{Projection Block}
    \label{fig:supp_details_proj}
  \end{subfigure}

   \caption{Network details for our shared Token / Confidence Head and Projection Block.}
   \label{fig:supp_details}
\end{figure*}

\section{Additional Implementation Details}
\label{sec:imple_details}

\subsection{\bf Training Details of CGD}
\cref{fig:supp_cgd_training} provides details of generating confidence labels during training. Since teacher-forcing is applied during training, multiple heads can predict and supervise each token. For any given token, the first head’s prediction is generally considered more reliable. Therefore, the confidence label $\hat{c}_{k+i}^j$ for the $(k+i)$-th token ${t}_{k+i}^j$, generated by the $j$-th head ($j \in [2, n]$), is computed based on its consistency with the token ${t}_{k+i}^1$ from the first head. If absolute difference satisfies $|{t}_{k+i}^j - {t}_{k+i}^1| \leq \tau$,  the token ${t}_{k+i}^j$ is regarded as correct and $\hat{c}_{k+i}^j=1$, otherwise $\hat{c}_{k+i}^j=0$. \footnote{In the main paper, we omit the head index $j$ for simplicity. The token $\widetilde{t}_{k+i}$ mentioned in the main paper corresponds to ${t}_{k+i}^1$ in the supplementary material.}

\subsection{\bf Training Settings} 
Training employs the AdamW optimizer~\cite{loshchilov2017decoupled} with an initial learning rate of $0.001$ and a batch size of $64$. A multi-step linear learning rate scheduler is used. The loss weights $\lambda_h$ and $\lambda_c$ are set to $0.8$ and $0.5$, respectively.

For layout estimation on the ASE dataset~\cite{avetisyan2024SceneScript}, the NTP model is trained for $60$ epochs. Following~\cite{cai2024medusa,DBLP:journals/corr/abs-2504-04060}, the MTP model is initialized from the NTP model, excluding the projection blocks and confidence heads, and trained for $30$ epochs. On the Structured3D dataset~\cite{zheng2020structured3d}, models are initialized with weights pretrained on ASE~\cite{avetisyan2024SceneScript} and further trained for $600$ epochs. Finally, the SceneCAD model is fine‑tuned on SceneCAD~\cite{avetisyan2020scenecad} for $600$ epochs.

For object detection on ASE dataset~\cite{avetisyan2024SceneScript}, the NTP model is trained for $150$ epochs, while the MTP model is trained for $120$ epochs using the pretrained NTP model. On SceneCAD~\cite{avetisyan2020scenecad}, models are initialized with weights trained on ASE~\cite{avetisyan2024SceneScript} and trained for $600$ epochs.

\subsection{Network Architecture}

The architecture details of our shared Token / Confidence Head and Projection Block are presented in \cref{fig:supp_details}. The Token Head and Confidence Head have a similar architecture, including 3 linear layers and 2 ReLU layers. The Projection Block is built from 2 feed-forward blocks, following the  Transformer FFN design~\cite{vaswani2017attention}. Each block consists of 2 linear layers, 1 ReLU layer, 1 Layer Normalization (LN) layer, and shortcut connection between its input and output.

\section{Additional Related Works}
\label{sec:supp_related_works}
This section reviews related work on 3D layout estimation from point clouds.

Scan2BIM~\cite{DBLP:conf/iros/MuraliSOP17} detects walls with conventional geometric analysis algorithms, such as One-Point RANSAC Model fitting~\cite{DBLP:journals/cacm/FischlerB81}. FloorNet~\cite{DBLP:conf/eccv/LiuWF18} uses an Integer Programming formulation~\cite{DBLP:conf/iccv/Liu0KF17} to convert pixel-wise predictions on layout geometry and semantics into vector-graphics layout. SceneCAD~\cite{avetisyan2020scenecad} predicts layout planes through a bottom-up pipeline, which generates corners, edges, and planes hierarchically. HEAT~\cite{DBLP:conf/cvpr/ChenQF22} detects corners and classifies edge candidates between corners with Transformer~\cite{vaswani2017attention}. RoomFormer~\cite{yue2023connecting} uses Transformer~\cite{vaswani2017attention} to predict a set of room polygons. Its Transformer decoder uses two-level queries: room-level queries and corner-level queries. SceneScript~\cite{avetisyan2024SceneScript} and SpatialLM~\cite{mao2025spatiallm} reformulate layout estimation as a language modeling problem, achieving state-of-the-art accuracy with a simple and end-to-end manner. SceneScript~\cite{avetisyan2024SceneScript} designs task-specific structured language and trains a specialized small language decoder, while SpatialLM~\cite{mao2025spatiallm} uses a general large language model~\cite{DBLP:journals/corr/abs-2407-21783,DBLP:journals/corr/abs-2412-15115}. Our Fast SceneScript is designed to significantly accelerate inference compared to SceneScript~\cite{avetisyan2024SceneScript}, while preserving its accuracy.

\section{Dataset Splits for ASE}
\label{sec:data_split}
This section describes our data splits of the ASE dataset~\cite{avetisyan2024SceneScript}.

For layout estimation, the split is defined in \textit{ase\_split/layout\_train.txt}, \textit{ase\_split/layout\_val.txt}, and \textit{ase\_split/layout\_test.txt}.

For object detection, the split is defined in \textit{ase\_split/detection\_train.txt} and \textit{ase\_split/detection\_test.txt}.

\end{document}